# The Semantic Adjacency Criterion in Time Intervals Mining

Alexander Shknevsky, Yuval Shahar, Robert Moskovitch

Department of Software and Information Systems Engineering,
Ben-Gurion University, Beer-Sheva, Israel.

sheknabs, yshahar, robertmo@bgu.ac.il

**Abstract**

Frequent temporal patterns discovered in time-interval-based multivariate data, although syntactically correct, might be *non-transparent*: For some pattern instances, there might exist intervals for the same entity that *contradict* the pattern's usual meaning. We conjecture that non-transparent patterns are also less useful as classification or prediction features.

We propose a new pruning constraint during a frequent temporal-pattern discovery process, the *Semantic Adjacency Criterion [SAC]*, which exploits domain knowledge to filter out patterns that contain potentially semantically contradictory components. We have defined three SAC versions, and tested their effect in three medical domains. We embedded these criteria in a frequent-temporal-pattern discovery framework.

Previously, we had informally presented the SAC principle and showed that using it to prune patterns enhances the repeatability of their discovery in the same clinical domain. Here, we define formally the semantics of three SAC variations, and compare the use of the set of pruned patterns to the use of the complete set of discovered patterns, as features for classification and prediction tasks in three different medical domains. We induced four classifiers for each task, using four machine-learning methods: Random Forests, Naïve Bayes, SVM, and Logistic Regression. The features were frequent temporal patterns discovered in each data set.

SAC-based temporal pattern-discovery reduced by up to 97% the *number* of discovered patterns and by up to 98% the discovery *runtime*. But *the classification and prediction performance of the reduced SAC-based pattern-based features set*, was *as good* as when using the complete set.

Using SAC can significantly reduce the number of discovered frequent interval-based temporal patterns, and the corresponding computational effort, without losing classification or prediction performance.

**Keywords**: Temporal Data Mining, Time Intervals Mining, Semantics, Frequent Pattern Mining, Classification, Prediction.

# 1. Introduction

This paper deals with the increasingly important topic of the discovery of frequent temporal patterns, given, as input, a set of *symbolic time intervals*, i.e., time periods over which hold one or more propositions, such as, in the medical domain, "The dose of the medication was High" or "The blood pressure was Low", and the temporal relationships among these periods. The discovered temporal patterns can then be exploited for clustering, classification, and prediction.

Analyzing time-oriented, multivariate clinical data enables researchers to discover new temporal knowledge and gain understanding regarding the temporal behavior and temporal associations of these data [Batal et al., 2012b; Klimov et al., 2015; Moskovitch et al., 2015; Sacchi et al., 2015]. The main methods for the discovery of new knowledge in longitudinal multivariate data include several *Temporal Data Mining* (TDM) approaches. Unlike most TDM methods, which typically focus mainly on the analysis of the *raw*, *time-stamped* data, the use of *symbolic time intervals* can reduce inherent random noise in the data, avoid problems resulting from different sampling frequencies and at various temporal granularities, and often alleviate the problem of missing data [Harel and Moskovitch, 2021; Lee et al., 2020; Lin et al., 2003; Mörchen and Ultsch, 2005; Mordvanyuk et al., 2020; Moskovitch and Shahar, 2015a; Rebane et al., 2020].

Thus, to significantly enhance the capabilities for analysis of time-stamped data, a preprocessing step of meaningful summarization and interpretation of the time-stamped raw data (e.g., a series of hemoglobin values) into a set of interval-based abstractions or symbolic time intervals (e.g., periods of moderate anemia), known as *temporal abstractions*, can be used [Batal et al., 2009; Moskovitch and Shahar, 2015c; Sacchi et al., 2007; Shahar, 1997; Verduijn et al., 2007]. The resulting interval-based summary can have multiple uses, such as to create natural-language free-text summaries of clinical data [Goldstein and Shahar, 2016], to visualize the data of individual patients [Martins et al., 2008; Shahar et al., 2006], or to interactively explore associations among the time-oriented data and their abstractions [Klimov et al., 2010].

Once a set of interval-based abstractions of the time-stamped raw data exists, a set of [sufficiently] frequent temporal patterns, incorporating these symbolic time intervals as components, can be discovered. We refer to these patterns as *time-interval-relation patterns* (*TIRPs*) [Moskovitch and Shahar, 2015a]. Within TIRPs, all of Allen's seven basic temporal relations and their respective inverse relations [Allen, 1983] might hold, such as Before.

Recently, time-interval patterns have been increasingly used as features to classify multivariate temporal data [Batal et al., 2012b; Dvir et al., 2020; Fradkin and Mörchen, 2015; Itzhak et al., 2020; Moskovitch et al., 2015; Moskovitch and Shahar, 2015b; Novitski et al., 2020; Patel et al., 2008]. Using that approach, the [sufficiently] frequent interval-based temporal patterns are used as the base features to induce a classifier. Furthermore, we have shown, in an earlier study, that frequent TIRPs can be consistently discovered, and in similar proportions, in different subsets of the same data set, within three different medical domains, especially as the minimal threshold for frequency is raised, thus increasing their value for potential classification and prediction tasks [Shknevsky et al., 2017]. This repeated discovery suggests that discovered TIRPs might indeed be good candidates for use as classification or prediction features.

However, our work in multiple clinical domains had suggested that many of the discovered frequent temporal patterns, although correct from the purely *syntactic* aspect, do not conform to the basic *semantics* of medical experts, who often assume a certain type of temporal adjacency among the temporal-pattern's components. Thus, such patterns are not *transparent* characterizations of the data. This semantic temporal adjacency, which domain experts seem to implicitly assume, means that no instance of the pattern includes any [additional] intermediate intervals within the scope of the pattern, which might contradict a potential interpretation of causality, or at least direct temporal association, among the pattern's components. We refer to this constraint as the *Semantic Adjacency Criterion* (SAC). In the current paper, we shall formally define it, explore its variations and semantics, and demonstrate its significance for TDM.

An example of the use of the SAC is the discovery of the following frequent pattern: <"A period during which a High dose of the medication" occurs *Before* "A period during which the Hemoglobin level was Low"> (Figure 1). A domain expert might assume that perhaps there is a causal association between the two symbolic intervals, since they frequently seem to be found together in this specific order. Perhaps administering a High dose of the medication might be associated with a Low Hemoglobin level?

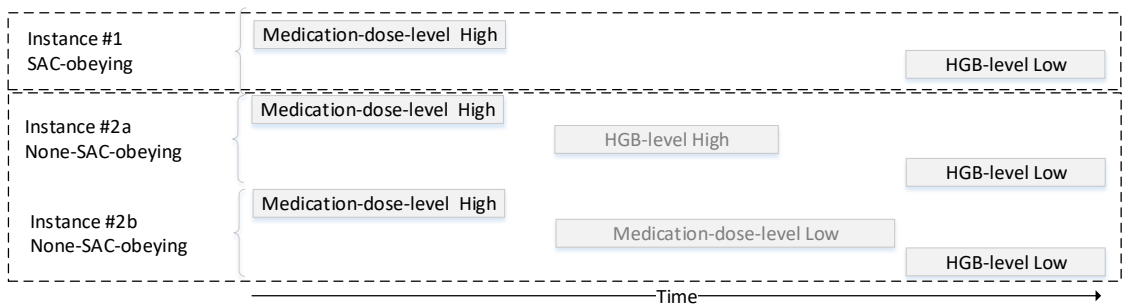

**Figure 1** - Three syntactically equal instances of the same interval-based temporal pattern, which includes the symbolic intervals, "<Medication-dose-level = High> Before <Hemoglobin [HGB]-level = Low>". Instance #1 describes a situation in which the two intervals are adjacent, and no contradicting value exists between them, and thus preserves semantic transparency. Instance #2a describes a situation in which the two intervals are *not* semantically adjacent, since there is an unexpected (from the point of view of the expert) High hemoglobin-level value between them that contradicts the pattern's semantics; Instance #2b, similarly, contains an unexpected medication-dose level (Low) between the two symbolic intervals. Modified from [Shknevsky et al., 2017].

However, what would the expert say if they will find that between these two symbolic intervals, there often exists, within the patient's original longitudinal record, an additional interval, during which a *High* level of Hemoglobin exists (Instance #2a in Figure 1)? Alternatively, what if the expert finds one or more instances of the pattern in which, between the two components defining it, there exists in the patient's record an interval during which dose of the medication was actually *Low* (Instance #2b in Figure 1)?

Although technically, the original temporal relation still holds, its significance now, from a medical expert's point of view, might change considerably. That would be the case whether the discovered frequent pattern is used for human explanatory purposes, or for succinctly summarizing the data, or, as we shall examine in this study, for machine-learning purposes.

As we formally define in the Methods Section, we distinguish a TIRP (an abstract pattern with certain temporal qualitative constraints) from its TIRP *Instances*, which are found in the longitudinal records that are being analyzed. A TIRP is *frequent* if the proportion of the records within which its TIRP instances are discovered is higher than some threshold. However, within the temporal scope of some of the instances of frequent TIRPs, there might exist, in some of the patients' longitudinal records, additional symbolic intervals (which are not part of that instance), as shown in Figure 1, which seem to contradict the TIRP's intuitive semantics.

Note that experts, especially medical experts, often expect a meaningful frequent temporal pattern to convey some potential causal relationship, such as a High dose of a medication reducing the level of Hemoglobin, in the case of the temporal pattern depicted in Figure 1, Instance #1. The fact that the true state of affairs is such that it rules that possibility out, as in the case of Instance #2a, since in the patient's record there exists a High-Hemoglobin period after the administration of the medication, would not be expected by a clinician when hearing the description of the frequently discovered pattern as "<Medication-dose-level = High> Before <Hemoglobin [HGB]-level = Low>". Nor would they expect that there might be an additional episode of medication administration, but with a Low dose, before the High Hemoglobin value. Given that description, from the point of view of a clinician, Instance #2a and Instance #2b lack *semantic coherence*.

In the current study, we are *not* exploring the purely *psychological* issue of the potential lack of transparency, to medical experts, of different temporal-pattern semantics. (Although such a lack might considerably reduce, for example, the patterns' explanatory value, their data-summarization value, and the efficacy of the experts in suggesting additional patterns to explore). What we *do* conjecture in this study concerns a purely quantitative, *functional* issue: We believe that such "*semantically incoherent*" patterns, besides being potentially less transparent to the *medical experts*, might also be less useful, and perhaps even unnecessary, as classification and prediction features for a *machine-learning process*, precisely due to their potential lack of semantic coherence. At the same time, discovering such redundant, "semantically incoherent" temporal patterns might require significant effort during the discovery time, as well as during classifier-induction time, without enhancing the accuracy of the resultant classifiers.

Of course, what precisely is and is not *semantically coherent* within a complete, multiple-interval temporal pattern needs to be carefully and formally defined, as we do in Section 3. We shall see that in fact, several options exist for exploiting the basic semantic intuition demonstrated in Figure 1, depending on which constraints exactly must hold in such a temporal interval triad, so as to comply with the notion of semantic coherence.

Although several of the earlier studies have noticed a potential redundancy during pattern discovery (in particular, the discovery of patterns containing repeating symbolic intervals as components, such as discovery of the pattern AAB in addition to the discovery of the pattern AB), or even considered patterns characterized by the *absence* of certain symbols [Höppner and Peter, 2014] they have only considered the issue from a purely *computational* point of view (i.e., the complexity of the temporal-pattern discovery process) [Batal et al., 2012a; Moskovitch and Shahar, 2015a]. For example, no attention was paid to the relationship between pattern components denoted by *different* symbols, each of which represents a different proposition, which, however, in a medical domain's ontology, might in fact represent different values of the *same* concept. For example, both of the symbols "*Low blood pressure*" and "*Hypertension*" (i.e., High or Very-High values of the blood pressure) are propositions that assign different values to the same concept, namely, the concept that denotes the abstraction of the raw-data Blood Pressure measurement concept into a discrete symbolic value. In contrast to these studies, in the current study, we shall refer to that potential problem from a *semantic* point of view (i.e., the potential *meaning* of the discovered pattern and of each of its components), as well as from a *functional* point of view (i.e., the implications for the *effectiveness* of the classification).

In an earlier study [Shknevsky et al., 2017], which had focused on the consistent discovery of sets of temporal patterns in various types of clinical databases, we had introduced the SAC principle in a highly informal fashion, and showed that using it to prune patterns enhances the repeatability of discovering the same temporal patterns in similar proportions in different patient groups within the same clinical domain. *In the current study, for the first time, we define, exemplify, and explain in detail the SAC principle's formal semantics for the three SAC variations that we are using, and then focus, through a series of quantitative machine-learning experiments, on its significant computational implications for classification and prediction.* We demonstrate that significance by comparing the use of the complete set of discovered temporal patterns, versus the use of only the set of pruned patterns, as features for several classification and prediction tasks in three different medical domains (oncology, hepatitis, diabetes), using, each time, four different machine learning approaches (Random Forests, Naïve Bayes, SVM, and Logistic Regression), all using the same TIRPs, and only TIRPs, as their basic features.

Thus, in the current study, we have explored the application of *semantic considerations to symbolic time-intervals mining, and to classification and prediction tasks, in medical domains*. The current study significantly extends our preliminary study of the potential value of semantic considerations in symbolic time intervals mining [Shknevsky et al., 2014]. It also complements our investigation of the consistency of discovering the same temporal patterns in similar rates in similar patient populations [Shknevsky et al., 2017]. *Besides formally introducing and explaining the SAC principle, the current study examines for the first time its exploitation for considerably reducing the number of temporal-pattern features used by multiple types of machine-learning methods, and the time needed to discover this high reduced set, without any loss of classification of prediction performance.*

As we shall see, the use of domain-specific semantics (which is explained in detail in Section 3.3) can constrain the discovery of temporal patterns in symbolic time intervals data to only those patterns that include certain semantically meaningful relations amongst the symbolic time intervals of which they are composed, in the sense of not violating certain semantic constraints that we have formally defined. Note that *no new temporal patterns are discovered*; rather, a large number of candidate patterns are *pruned [filtered] out* during the discovery process. Thus, our main contribution in this study, besides introducing, for the first time, a highly detailed and formal definition of the SAC principle and several of its variations, is the rigorous evaluation for classification and prediction purposes of a new pruning constraint for mining time intervals, the *Semantic Adjacency Criterion* (SAC). In fact, we have defined and explored three different versions of the SAC criterion.

Consequentially, *our core, double-pronged hypothesis in this study is that:*
*(a) It is more efficient, during the temporal data mining process, to discover only semantically coherent patterns [coherent in a sense that we shall formally define]; But nevertheless,*

*(b) Imposing such semantic constraints, leading to the discovery of a significantly smaller set of patterns, will not cause any harm with respect to the performance of the discovered patterns as features for classification or prediction purposes, and might even enhance that performance.*

As our current study demonstrates, using any of the SAC versions' results in the discovery of temporal patterns whose overall *cardinality,* as well as *time needed for discovery,* are *smaller by at least an order of magnitude* than the respective resulting cardinality and required running time, when not using the SAC constraint; but that whose *value to classification and prediction tasks* that use the discovered patterns as features is *at least as good as the original full set of discoverable patterns*.

The outline of the paper is the following: Section 2 provides the necessary background for the rest of the paper. Section 3 describes our computational framework, including the use of temporal abstractions, the discovery of TIRPs, and the formal definition of the three SAC versions. Section 4 describes the evaluation and the experiments we performed to assess the effect of using the three SAC versions to discover frequent temporal patterns and exploit them as features for classification and prediction purposes, using four different classifier-induction methods within each of three different clinical domains, while Section 5 describes the results of our empirical evaluation. Section 6 discusses the results and presents the main conclusions of this study.

## 2. Background

In this section we briefly present the background topics that are most relevant for later presenting our methodology in detail, including: Semantics of Symbolic Time Intervals, Time-Intervals Mining, and Classification using Temporal Patterns as features.

### 2.1 The Structure and Semantics of Symbolic Time Intervals

As the reader might have already gleaned from the examples mentioned in the previous section, the *symbols* that may hold on symbolic time intervals usually denote the combination of a *concept* and its *value*. A concept might represent *raw* data (e.g., a Blood-Pressure measurement) or an event (e.g., administration of the medication Insulin). The *values* of such concepts are often numeric, such as "90 mmHg" for a blood pressure measurement, or "2 units" for an Insulin administration. Other raw-data concepts, whose default value is "True", include events such as a total-hip-replacement surgery. A concept might also denote, however, a more abstract interpretation of the raw-data concept, such as the [discretized] level of the Blood Pressure raw-data concept, or the assessment of the dose of administered Insulin. In that case, the respective values might be 'High' or "Low dose" or "Very High".

Symbols that contain Abstract-concepts might also be the result of a *temporal-abstraction* process, in which a series of raw time points were transformed into one or more symbolic time intervals, such as the concept "The trend of the Blood-Pressure measurements" with the value "Decreasing".

Several types of abstract concepts, and in particular temporal abstractions, exist. To be clear and consistent throughout this paper, we use a simple, well-known temporal-abstraction ontology, which has been deployed in multiple application domains, which is the ontology used by the *Knowledge-Based Temporal Abstraction* (KBTA) method [Shahar, 1997]. *However, both the SAC principle and the results of using it are quite generic, and do not depend on the use of any particular temporal-abstraction ontology, nor on any particular methodology for generating the symbolic time intervals.*

An abstract concept might denote a *State* abstraction, i.e., a classification of the value(s) of one or more raw-data concepts into a set of values of a single abstract concept, using a set of cutoff values. An example is the abstraction of the Hemoglobin-level value into several states, such as Normal_Hemoglobin or Moderate_Anemia. Similarly, the raw-data Height and Weight concepts might be jointly abstracted, using a simple arithmetical function (Weight/Height$^2$), into the abstract concept of a *body mass index* (BMI). The BMI concept, in turn, might be further abstracted, using simple cutoff values, into the "BMI state" abstract concept, which might have values such as "underweight", "Normal_weight", "overweight", or "obese". A state abstraction can be performed using cutoff values that are provided by a domain expert [Shahar, 1997]. Alternatively, the cutoff values might be derived directly from the raw data series [Lin et al., 2003; Mörchen and Ultsch, 2005; Moskovitch and Shahar, 2015c].

Other types of abstractions can be generated from raw time-stamped data, such as *Gradient* (e.g., *Decreasing), Rate* (e.g., *Fast*), *Trend* (e.g., *Stable*) abstractions, and even whole patterns represented as a single symbolic interval [Shahar, 1997; Sheetrit et al., 2019]. Our framework caters equally for all of them, but for simplicity's sake, we shall focus in our discussion only on State abstractions.

Temporal abstractions are usually formed within a *context*, or a state of affairs, such as being a male or a female, an infant, or being under the influence of an Insulin injection [Shahar, 1997]. Thus, the knowledge necessary for correctly forming abstractions from raw data is context-sensitive, and the concepts, implicitly or explicitly, might include that context (e.g., "the State of the Hemoglobin-value of a young woman"). For our purposes in the current paper, we shall assume that the context is a part of the concept. Several approaches to the abstraction of time-oriented raw data into symbolic intervals exist. The KBTA methodology uses domain-specific and context-sensitive *classification knowledge* to generate the abstractions from raw concept values, and applies *temporal interpolation knowledge* to bridge gaps between time points and time intervals and join them into longer [symbolic] time intervals. It was applied within multiple domains and to different tasks, such as within the domains of medicine, biology, information security, or traffic control, and to the tasks of summarization, visualization, exploration, classification, and prediction [Klimov et al., 2015; Martins et al., 2008; Shabtai et al., 2010; Shahar et al., 2006; Shahar and Molina, 1998; Shahar and Musen, 1993].

However, when no suitable domain knowledge exists, various data driven discretization methods exist [García et al., 2013], which typically focus on finding cut-off values (using various heuristics) for discretizing continuous data. Such methods include, for example, the *Equal Width Discretization* (EWD) method, which has been demonstrated to often be sufficient for purposes of classification or prediction [Azulay et al., 2007; Moskovitch and Shahar, 2015b]; the SAX method for discretization of time series [Lin et al., 2003]; and the *Temporal Discretization for Classification* (TD4C) method [Moskovitch and Shahar, 2015c], a discretization method that is specifically geared for the classification task. The TD4C method learns the state-abstraction cutoff values, which best separate the instances belonging to the predicted classes with respect to their differing distribution of abstraction values over time. The inter-distribution distance measure is either Cosine, Entropy, or the Kullback–Leibler measure. The TD4C method outperformed the EWD and SAX discretization methods, for the purpose of classification, in several different medical domains [Moskovitch and Shahar, 2015c].

As we shall now see, one can mine symbolic time intervals to discover frequently occurring temporal patterns in the data of multiple subjects.

## 2.2 Mining Symbolic Time Intervals

Typically, time interval mining methods use some subset or variation of Allen's temporal relations [Allen, 1983]. Allen defined 13 temporal relations, based on 7 relations (*before, meets, overlaps, finished-by, contains, start-by, equal*) and their inverses (note that the inverse of *equal* is *equal*). Another option, which in this study we shall investigate as well, is the one of using only three abstract temporal relations, two of which are defined by a disjunction of Allen's relations [Moskovitch and Shahar, 2015a]: BEFORE, which is the disjunction of {*before, meets*}, OVERLAPS, which is the usual *overlaps*, and CONTAINS, which represents the disjunction of {*finished-by, contains, started-by, equal*}.

Höppner introduced a method to mine rules in symbolic time interval sequences using Allen's temporal relations, using a non-ambiguous representation through a conjunction of the pairwise temporal relations among the symbolic time intervals [Höppner, 2001]. This time intervals patterns definition was later used to discover patterns more efficiently by several groups [Moskovitch and Shahar, 2015a; Papapetrou et al., 2005; Patel et al., 2008; Winarko and Roddick, 2005]. Other researchers used additional abstract relations [Sacchi et al., 2007] or other types of temporal relations, such as coinciding [Moerchen, 2006]. Interval mining, and in particular symbolic (abstract) interval mining, has gathered significant traction, and multiple recent studies have demonstrated various improvements on mining interval-based patterns [Harel and Moskovitch, 2021; Lee et al., 2020; Mordvanyuk et al., 2020; Rebane et al., 2020].

Note that the task of *interval-based* temporal data mining is quite different from *sequential mining* (see Section 2.3), which, as its name would suggest, is purely sequential, and usually focuses on point-based episodes. The main task in the case of the multiple algorithms for mining patterns based on time intervals is to mine frequent patterns of repeating temporal relations among multiple time intervals. That task includes the determination of temporal relations such as *contains*, *overlaps*, and *finishes*, in addition to the standard *before* and *after* and *equals*.

## 2.3 Classification and Prediction Based on Temporal Patterns

The field of sequential data mining and its use for TDM tasks has been explored in various ways, e.g., through the sequence classification task [Xing et al., 2010], or through sequence and motifs mining to extract features for classification [Buza and Schmidt-Thieme, 2010; Ferreira and Azevedo, 2005; Lesh et al., 1999]. However, its focus is on time-stamped events, and essentially only on the before relation, and so is not appropriate for the types of domains and tasks in which we are interested. We are interested in the more general nature of multivariate, time-stamped and interval-based data. Quite simultaneously, several groups proposed using TIRPs as features for classifying multivariate time series [Batal et al., 2012b, 2012a; Moskovitch et al., 2009; Patel et al., 2008], which were followed by recent studies exploiting temporal patterns as features [Dvir et al., 2020; Itzhak et al., 2020; Moskovitch and Shahar, 2015a, 2015b; Novitski et al., 2020].

Finally, the *distribution* of frequent TIRPs was effectively exploited for prediction, using the concept of *Temporal Probabilistic proFiles* (TPFs) and the distance between two frequent-pattern distributions as the major similarity measure for classification and prediction, surpassing in performance, in the case of sepsis prediction in the intensive-care unit, both the use of the pure temporal patterns as features and the use of deep learning algorithms learning from only the multivariate, time-oriented raw data [Sheetrit et al., 2019].

Interestingly, all of the above studies reported the use of temporal abstraction and the use of TIRPs for classification applied to biomedical data. Patel et al. proposed IEClassifier to classify Hepatitis patients using TIRPs [Patel et al., 2008]. Batal et al. performed knowledge-based temporal abstraction, but used only two relations: before and co-occur, which is a specific case of an a priori sequential mining algorithm called STF-Mine [Batal et al., 2012b]; Several studies had shown the advantages of using TIRPs over a-temporal representation in classifying multivariate temporal data [Batal et al., 2012a, 2012b; Moskovitch et al., 2015, 2016; Patel et al., 2008]. Recent studies introduced several heuristics to decrease the number of discovered patterns that still maintain the same level of accuracy [Fradkin and Mörchen, 2015; Shknevsky et al., 2014].

Moskovitch and Shahar presented KarmaLegoSification – a framework for classification of multivariate time series via temporal abstraction and time intervals mining [Moskovitch and Shahar, 2015b]. Two new metrics were defined: *horizontal support,* which represents the number of TIRP instances discovered for a specific entity, and *mean duration*, which measures the average time length of the supporting TIRP instances. Both new feature representation methods were shown to be superior to the default binary representation for classification. The use of the three abstract temporal relations (Section 2.2) was superior to the use of Allen's seven relations, and using knowledge-based State abstractions (see Section 2.1), when available, performed better for classification purposes than using EWD or SAX [Moskovitch and Shahar, 2015b]. However, in a later study, using the TD4C method (see Section 2.1) to create State abstractions outperformed even the use of a knowledge-based abstraction method for that purpose.

As we shall see when presenting our methods and results, we have used several insights from these preceding studies when assessing the value of our new semantic criteria for the purpose of significantly reducing the number of patterns, without reducing their classification performance.

## 3. Methods

We start by first defining the basic interval-based TDM terminology. We then describe briefly the high-level overview of the general TDM algorithm we chose to deploy for this study, before formally introducing the SAC criterion and its semantics and several versions.

## 3.1. The Time Intervals Mining Process

To formally define the problem of mining symbolic time intervals, and to better comprehend how the SAC constraint can be introduced into an interval-based frequent-pattern discovery algorithm, we present several basic definitions used by common TDM algorithms, such as by the KarmaLego frequent-pattern discovery algorithm [Moskovitch and Shahar, 2015a], which we are also using within the evaluation.

For the purpose of our current discussion, we shall use few of [Moskovitch and Shahar, 2015a]'s definitions.

We define a *symbolic time interval*, I = <s, e, sym>, as an ordered pair of time points, start time (*s*) and end time (*e*), and a symbol (*sym*), which represents one of the domain's symbols from the domain-specific set Ś.

A *non-ambiguous lexicographic TIRP P* = {Ï, Ṙ} is defined as a set Ï of *k* symbolic time intervals ($I_1...I_k$) ordered lexicographically by the start time, then (for identical start times) by the end time, then (if both are identical) by the symbol, and a set Ṙ of all of the pairwise temporal relations among each of the $(k^2-k)/2$ pairs of symbolic time intervals in Ï. Note that *each TIRP is an abstract temporal pattern* that represents a *class*, or a *set*, of specific *instances* of that TIRP. (The input interval-based database is also ordered lexicographically for each entity, such as each patient.)

The goal of discovering frequent TIRPs is to discover [sufficiently] frequently occurring abstract patterns within the instances of the given database; each pattern instance found in the database is a *TIRP instance*.

The fact that the database is ordered lexicographically for each entity enables us to use only seven of Allen's temporal relations (or even the 3 abstract relations) as defined in section 2.2. Figure 2 presents a typical TIRP, represented as a half-matrix of temporal relations.

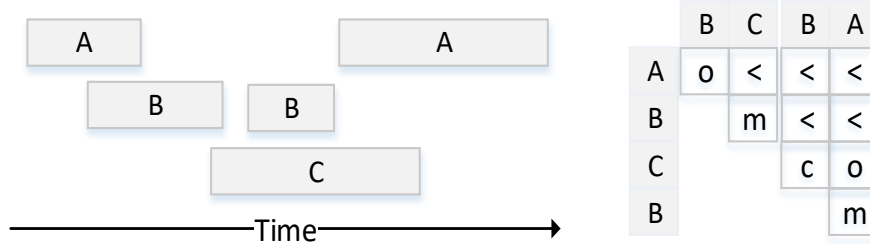

**Figure 2** - An example of a TIRP representation, containing 5 symbolic time intervals, and all of their pair-wise temporal relations. Modified from [Moskovitch and Shahar, 2015a].

The *vertical support* for a TIRP is defined as follows: given an input database of |E| distinct entities (e.g., different patients), each represented as a set of symbolic time intervals (in which one or more symbols might repeat over different symbolic time intervals), the *vertical support* of a TIRP *P* that was discovered in the input data is denoted by the cardinality of the set $E^P$ of distinct entities (within which at least one instance of *P* was discovered) divided by the cardinality of |E|.

When a TIRP has a vertical support above a given minimal predefined threshold *min_ver_sup*, it is referred to as *frequent*. Furthermore, the *horizontal support* of a TIRP *P* for an entity e (e.g., a single patient's record), hor_sup(P, e), is the number of instances of P found in e. Accordingly, we also define the mean duration of the supporting instances of the same TIRP P within an entity e as the average time of the instances of P within a specific entity e, from its earliest start-time to its last end-time. Thus, given a minimum vertical support *min_ver_sup*, the goal of the mining task is to find the complete set of the frequent TIRPs, including all of their supporting instances vertically and horizontally [Moskovitch and Shahar, 2015b].

The KarmaLego algorithm [Moskovitch and Shahar, 2015a], which we chose to use in this study's evaluation to discover frequent temporal patterns due to its proven efficiency comparing it to several of the best known alternatives (and it is provably complete in its discovery of all frequent TIRPs [Moskovitch and Shahar, 2015b]), consists of two main phases. The first phase is called Karma, in which all of the frequent 2-sized TIRPs, having two symbolic time intervals $I^1$ and $I^2$ and a temporal relation *r* among them, are discovered and indexed. In the second phase, called Lego, a recursive process extends the frequent 2-sized TIRPs, referred to as $T^2$, through efficient candidate generation, into a tree of longer frequent TIRPs consisting of conjunctions of the 2-sized TIRPs that were discovered in the Karma phase. The final output is an enumeration tree of all the frequent TIRPs discovered in the given database.

## 3.2. Adding Semantic Considerations to Time Intervals Mining

As explained in Section 1, there is a potential drawback inherent in the interval-based pattern mining task. Many of the discovered patterns are *syntactically* true, but *semantically* misleading. Addressing this problem requires the addition of semantic considerations to the time intervals mining task, as we proposed in a preliminary study [Shknevsky et al., 2014].

## 3.3. The Semantic Adjacency Criterion

Since frequent TIRPs mining algorithms generate *all* of the feasible TIRP candidates and search for them in the data, certain discovered TIRP instances (e.g., a period of High-dose medication of a certain type, is followed by a period of Low blood pressure), although *syntactically* accurate, i.e., corresponding to a TIRP formal definition, might not represent in a transparent fashion the common-sense semantics that a domain expert might assign to the real data (see Figure 1). Thus, many of the discovered TIRPs might not be sufficiently transparent to the expert. We shall now explore this observation in depth.

Recall that a symbol, and in particular a symbol that holds during the duration of a symbolic time interval, is composed of a [raw or abstract] concept and its value (Section 3.1). In Section 1, we presented a frequent TIRP that might be discovered (Figure 1), "<Medication-dose = High> occurs before <HGB-level = Low>". The TIRP seems to imply that administering a medication at a High dose, is often [temporally] followed by a Low value of the Hemoglobin-value abstract concept (which is a State abstraction of the HGB-value raw-data concept; see Section 2.1). Such an association is not necessarily causal, of course, but it certainly might be, and justifies additional exploration.

To facilitate our discussion, for each symbol we will refer to its two components, the *concept* and its *value,* following, purely for consistency and clarity reasons, the KBTA theory's nomenclature [Shahar, 1997] (see Section 2.1). In the example we just discussed, the *state* abstractions "Medication-dose-level" and "HGB-level" are the [abstract] *concepts*, and "High" or "Low" are their *values*. A concept can only have one value at any point in time; different values during the same time are considered mutually exclusive and therefore *contradictory*. However, the TIRP shown in abstract fashion in Figure 2 might in fact include, within its supporting instances group [for a given database] that defines its *vertical* (or *horizontal*) *support* (see section 3.1), instances that include, somewhere within their overall temporal scope (although not at the same time), symbolic time intervals that represent values that are semantically *contradicting* (see explanation below) to those appearing in the formal TIRP definition. Two such cases were shown [as the grayed-out symbolic intervals of Instances #2a and #2b] in Figure 1. Perhaps administering a High dose of the medication might be associated with a Low Hemoglobin level?*Semantic Contradictions* are instances in which, between two of the TIRP's symbols, there is a symbol, composed of a concept and a value, such as, in this case, "HGB-level = High" (or "Medication-dose-level = Low"), in which the *concept* (which implicitly includes its abstraction type, such as State or Gradient, and its context, such as gender = Female), is *identical* to the concept of either of these two symbols, but its *value* is *different*. Either of these contradictory associations (with the first or the second symbols) might change, and in this case even reverse, the semantic meaning of the original temporal association, since it now seems that a *High* medication-dose level might be actually associated with a *High* Hemoglobin value (or conversely, that a *Low* medication-dose level is associated with *Low* hemoglobin level).

Note that even if the meaning of the intermediate symbol does not directly contradict the meaning of either of the temporal relation's symbolic time intervals, for example if both its concept name and abstraction type and its concept's value are the *same* as the concept name, type, and value of one of the two symbols, it might nevertheless change the overall pattern's semantics, or might simply be redundant. Thus, we might not wish to encounter even a *copy* of one of these two symbolic intervals between them.

It is important to note at this point that most time intervals mining algorithms, including KarmaLego, do not consider the semantics of the deeper structure of the symbols that hold over the symbolic time intervals, and thus view them as a single, non-decomposable symbol *sym*. However, as we have explained in Section 2.1, these symbols are in fact typically composed of a *concept* of some type and its *value* (e.g., the abstract concept "HGB-level" and its value "High"). Furthermore, using Shahar's KBTA ontology [Shahar, 1997] (see Section 2.1), an abstract concept would include also the *abstraction type* (e.g., *State*) and usually also a *context* (e.g., gender=Female). For example, the following predicate is an example of an abstraction that might hold for some patient: "The *State* of the *HGB*-value, in the context of a *Female, is High*". (Note that the definition of the value *High* might vary in different contexts.). We often refer to the full representation of the concept embodied in each symbol (which typically also includes an abstraction type and a context) as its **semantic type.** In the case mentioned above, the semantic type would be "The *State* of the *HGB*-value, in the context of a *Female*". The **value** of that semantic type (i.e., of the symbol's concept), would be "*High*".

**Definition 1.** Symbolic time intervals $I^i$ and $I^j$ are of the *same semantic type*, if each of the two symbols that hold over $I^i$ and $I^j$ represents some value for the *same concept* (which, as we recall, might include also a temporal abstraction type and a context). We denote that equivalence by the notation $sem\_type(I^i) = sem\_type(I^j)$. For example, the High value for the blood-pressure-level concept, and the Low value for the blood-pressure-level concept.

The semantic types (as defined above) of the symbols that might hold over all symbolic time intervals need to be pre-defined and used throughout the TIRPs discovery process. Such semantic types and the set of the allowed values for each abstraction of each concept in each context are a part of each domain's temporal-abstraction ontology [Shahar, 1997] (see Section 2.1). Thus, there might be, for example, exactly five values for the state abstraction of blood-glucose values in the context of a patient who has diabetes, ranging from hypoglycemia to hyperglycemia. Our intent is to discover only TIRPs in which adjacent symbolic time intervals are semantically coherent, i.e., their symbols are composed of concepts and values that fulfil a new, semantically oriented criterion, the *Semantic Adjacency Criterion (SAC)*.

**A Semi-formal Definition**: *The SAC guarantees that between two symbolic time intervals within a TIRP, there can exist no other symbolic time interval of the same semantic type as either of the two symbolic time intervals, nor any part of such an interval*.

In particular, over such an intermediate "forbidden" symbolic interval there cannot hold a symbol whose *semantic type*, i.e., its conceptual aspect (e.g., Hemoglobin-State in a Female, or the State of the Medication-dose), is the *same* as the semantic type of one of the two symbolic time intervals, but with a *different value* (e.g., a LOW value of the Hemoglobin State abstraction, instead of a HIGH value; or a LOW value of the Medication-dose State abstraction). Our motivation in defining and using the SAC is that symbolic time intervals that appear between a pair of two other symbolic time intervals, but are of the same semantic type as that of one of the two symbolic intervals, and in particular, those that contradict the value of one of the symbols that hold over the two symbolic time intervals, are not easily understandable to a domain expert and thus, the discovered TIRP might not really represent the associations the expert expects to find in the data.

The expert will be notified that the temporal data mining algorithm discovered a frequent relation such as "A before B", as in the case of "a *high* medication-dose before a *low* value of the hemoglobin level", not realizing that there might be another symbolic time interval between them that contains a concept of the same *type* as that of A or B (i.e., a medication dose, or a hemoglobin level), but with a similar, or even a different *value*.

Instances of such a potentially contradictory intermediate symbolic time interval might also interfere with the learning (training) phase of an algorithm that induces a classifier, and reduce its classification power, which relies on features that are TIRPs that were discovered while using the (potentially deceptive) temporal relation between that pair of symbolic time intervals. The reason is that the real meaning of that temporal relation might change in a radical fashion, depending on what other symbolic intervals exist between the two members of the pair, for some of the TIRP's supporting instances.

Thus, detecting in the patient's record an instance of Low blood pressure, and that two weeks ago she had taken a Low medication-dose of a medication that is similar to the medication that was taken yesterday at a High dose, i.e., a temporal pattern that syntactically also complies with the temporal relation of being *before* a current instance of Low blood pressure, might potentially mislead the classifier by erroneously suggesting to look for the temporal pattern "Low medication-dose before Low blood pressure" as a feature. However, a medical domain expert analyzing the same data set would usually consider as meaningful only a recent instance, often even only the *last* instance, of the medication administration before the blood pressure measurement, i.e., "<Medication-dose-level = High>  Before <Blood-Pressure-level = Low>" (which suggests that taking a High dose of the medication might have led to a Low blood pressure). At the very least, the expert would only consider the full sequence, "<Medication-dose-level = Low> Before <Medication-dose-level = High> Before <Blood-Pressure-level = Low>". She would usually be quite puzzled, or even misled, by the "discovery" of the pattern "<Medication-dose-level = Low> Before <Blood-Pressure-level = Low>".

We conjecture that using the SAC constraint, in addition to the discovery of only semantically meaningful patterns, will also significantly reduce the number of potential frequent TIRPs to consider, thus reducing the computational requirements of TIRPs discovery. The SAC constraint was inspired by the temporal interpolation mechanism from the KBTA methodology theory [Shahar, 1997, 1999]. The *temporal interpolation mechanism* uses, in each domain, a domain-specific *interpolation function* (found as part of that domain's temporal-abstraction ontology). The interpolation function is provided, as input, two symbolic time intervals, both of which hold similar temporal abstraction types (see Section 2.1) (e.g., Gradient) of the same concept (e.g., HGB level), such as two Increasing HGB-level periods, each lasting for two weeks, with a gap of one week between them, and that returns an abstraction, interpreted over an interval that joins the two intervals while bridging the gap between them, i.e., "five weeks of Increasing HGB level".

The temporal interpolation mechanism [Shahar, 1999] allows for a certain value-sensitive and context-sensitive maximal time gap to be bridged between the two symbolic time intervals. It also ascertains that the values of the symbols that hold over the symbolic time intervals within the gap, do not contradict in any way the values of the symbols that hold over the two symbolic intervals that are to be joined.

In the current study, we shall focus on only the *State* temporal abstraction (the one most commonly used in most TDM studies), although the methodology is equally relevant to all temporal-abstraction types. Furthermore, within the SAC principle, we shall focus only on the *Before* temporal relation between the two intervals of interest, which is the relation most relevant for our purposes. However, in our formal definitions and through our three SAC variants, we shall explore all of Allen's 13 temporal relations with respect to the relationship that *other* symbolic intervals in the same entity's database (e.g., the same patient's medical record) might have with the two intervals of interest.

Figure 3 closely examines the possible contradictions that might be hidden within a 2-sized TIRP, thus revealing several possible versions of the SAC. For example, in the simple case of sequential data mining, a TIRP can be treated as a sequence of symbols (considering the *before* and *meets* relations only). The symbol that holds over the intermediate symbolic time interval cannot represent the same concept (i.e., have the same semantic type) as one of the two symbolic intervals, even if it has the same value.

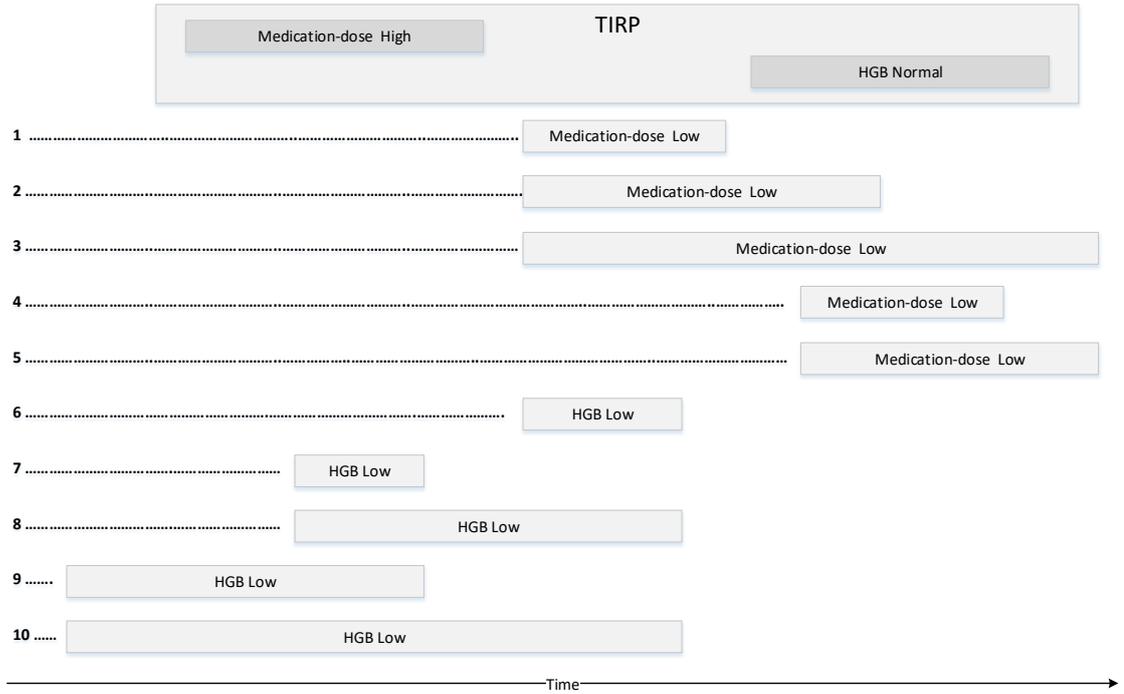

**Figure 3** - Example of possible contradicting instances within a 2-sized TIRP, when considering additional symbolic intervals that might exist in the same database, and the full range of temporal relations possible between two intervals. Cases 1, 2, 3, 6, 8, and 10 contradict the semantics of the TIRP defined above. Cases 4, 5, 7, and 9 appear outside the temporal relation gap within the TIRP and do *not* contradict it according to our current SAC definition

## 3.4. The Sequential Semantic Adjacency Criterion: A Formal Definition

Using this simple version of the SAC constraint, any two *successive* symbolic time intervals of each temporal relation pair within a TIRP, when the symbolic intervals of the TIRP are ordered *lexicographically* (see Section 3.1), and where the relation between them is *Before*, must be temporally adjacent, in the sense of our semi-formal definition. That is, no symbolic time interval whose symbol has a semantic type equal to one of them can exist *between* them. Note that the two symbols of the successive symbolic intervals themselves might include the same concept. This version of SAC is called the *Sequential Semantic Adjacency Criterion*.

**Definition 2.** The *Sequential Semantic Adjacency Criterion* (Sequential SAC, or SSAC) holds over a TIRP $P = \{\ddot{I}, \dot{R}\}$, where $\ddot{I} = \{I^1, I^2, \ldots, I^k\}$ and all of the TIRP's intervals are ordered lexicographically as defined above, iff:

$$\forall i, 1 < i < k-1: (I^i.e < I^{i+1}.s)$$
$$\rightarrow \nexists t: \left( (I^t.e > I^i.e) \land (I^t.s < I^{i+1}.s) \right)$$
$$\land \left( sem\_type(I^i) = sem\_type(I^t) \lor sem\_type(I^{i+1}) = sem\_type(I^t) \right)$$

(Note that $I^i$ is *before* $I^{i+1}$; relaxing this constraint and allowing for relations such as *Contains* or *Overlaps* leads to additional SAC versions that we introduce later.)

Definition 2 says, in effect, that given the TIRP *P*, there cannot exist in the database, for the same entity (e.g., the same patient), some interval $I^t$ with the same semantic type as $I^i$ or $I^{i+1}$ (where $I^i$ is before $I^{i+1}$ and $I^i$, $I^{i+1}$ appear successively in the lexicographic ordering), such that at least *some* of $I^t$ spans at least *some* of the gap between them. This is essentially the constraint that is exemplified graphically in detail by all of the semantically contradictory instances appearing in Figure 3, but the constraint is enforced only for *lexicographically successive* intervals.

However, note that considering only *successive* symbolic time intervals within the TIRP definition, using the lexicographic ordering, ignores the existence of the rest of Allen's temporal relations [Allen, 1983], and stays within the stricter limits of the sequential data mining approach. We shall demonstrate this observation with an example.

Figure 4 considers the TIRP "<Medication-dose = High> *before* <HGB = High> *before* <HGB = Normal>"; the relationship between the medication administration time interval and the third time interval is also *before*: "<Medication-dose> *before* <HGB = Normal>", but the two symbolic time intervals participating in this relation are not *successive* in the SSAC sense, since they do not follow each other in the lexicographic ordering.

Thus, in the Sequential SAC version, only the first two temporal relations of this particular TIRP's definition will be checked for the existence of any semantically contradictory (in the SAC sense) symbolic time interval instances between them. Thus, if the entity's data happen to include also a Low medication dose between the High medication dose and the High HGB level, that would be an SSAC contradiction, since these two intervals are successors in the TIRP's lexicographic listing. However, other semantically contradictory symbolic time interval instances might exist between the first and the third symbolic time intervals, which will *not* be checked using the SSAC. (For example, a Low medication dose within the dashed interval of Figure 4, i.e., contained within the span of the abstraction "<HGB = High>," would not contradict the relation "<Medication-dose = High> *before* <HGB = Normal>", since the components of that relation do *not* follow each other, i.e., are *not* successors, in the canonical lexicographic-ordering representation of this particular TIRP.)

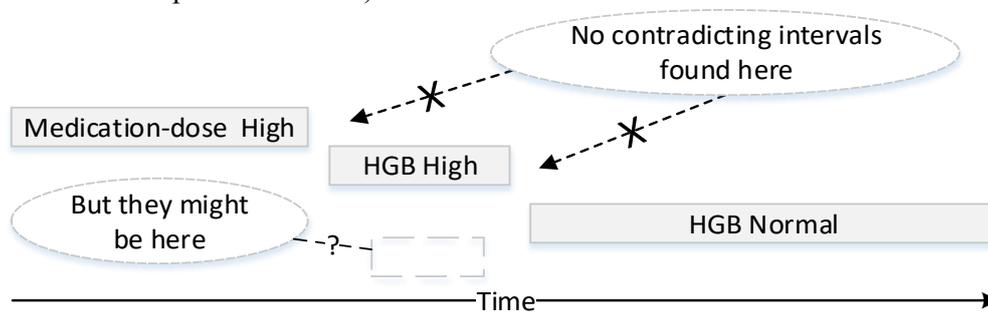

**Figure 4** - An example of the difference between the sequential version of SAC and other possible versions that do not consider only successive symbolic time intervals

Note two interesting insights regarding the SAC principle; First, this version of SAC does allow the discovery of what we refer to as "*Symbolic Gradient*" temporal patterns, e.g., Decreasing or Increasing values of the State abstractions of the same raw-data concept that hold over successive symbolic time intervals, such as a gradual decrease in the value of the Hemoglobin-State concept (see Figure 5). The reason is that the constraint will only be checked between every two successive symbolic time intervals, and these can be of the same type. Second, SSAC also allows the discovery of what we refer to as "Counting" temporal patterns. "*Counting*" *temporal patterns* are patterns that include some finite repetition of instances of a symbolic interval as part of the pattern. For example, the repeating of two or more successive Low Hemoglobin State abstraction values (see Figure 5). Such TIRPs might serve as useful features in certain domains.

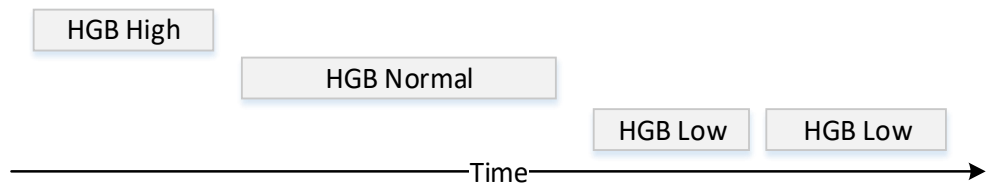

**Figure 5** - A possible legitimate (SSAC-obeying) TIRP that might be discovered by using the SSAC; the first three symbols represent a "Symbolic Gradient" temporal pattern of decreasing values of the Hemoglobin State abstractions, while the last two symbols present a "Counting" temporal pattern of two successive low hemoglobin tests

We have also defined two additional versions of SAC that capture slightly different semantics and enable different levels of expressivity of TIRPs that can be discovered.

## 3.5. The Conservative Semantic Adjacency Criterion: A Formal Definition

The first additional SAC version that we examine, considers *every* pair of symbolic time intervals within the TIRP definition. While the previous version considered only successive symbolic time intervals, we might want to consider *all* of the temporal relations within the TIRP's definition (thus probably reducing even more the potential vertical support for such a TIRP, and, correspondingly, the number of different TIRPs in the output).

Allowing the SAC constraint to hold over *all* of the TIRP's pairwise relations enables us to discover, in the input interval-based data, only true SAC-obeying TIRPs, in the most restrictive interpretation. Thus, unlike the SSAC, this version of SAC will *not* allow the discovery of any "Symbolic Gradient" temporal patterns or of any "Counting" temporal patterns at all (if the temporal pattern includes more than two intervals). Thus, for example, neither the TIRP displayed in Figure 5, nor any three components of it, would be considered as legitimate TIRPs according to this conservative constraint, which does not allow any potential semantic contradiction between *any* two intervals. We refer to this version of SAC as the *Conservative Semantic Adjacency Criterion*.

**Definition 3.** The *Conservative Semantic Adjacency Criterion* (Conservative SAC or CSAC) holds over a TIRP P = {Ï, Ṙ}, where Ï = $\{I^1, I^2, ..., I^k\}$ iff:

$$\forall i, j, \ 1 < i < j < k: (I^i.e < I^j.s)$$
$$\rightarrow \nexists t: \left((I^t.s < I^j.s) \wedge (I^t.e > I^i.e)\right)$$
$$\wedge \left(sem\_type(I^i) = sem\_type(I^t) \vee sem\_type(I^j) = sem\_type(I^t)\right)$$

Definition 3 says, in effect, that given the TIRP *P*, there cannot exist in the database, for the same entity, some interval $I^t$ with the same semantic type as $I^i$ or $I^{i+1}$ (where $I^i$ *completely precedes* $I^{i+1}$ in *some* way, directly or indirectly), such that at least *some* of $I^t$ spans at least *some* of the gap between them. This is again essentially the constraint that is exemplified graphically in detail by all of the semantically contradictory instances appearing in Figure 3; but $I^i$, $I^{i+1}$ are not necessarily *successors* in the TIRP's lexicographic ordering (unlike the case of SSAC).

## 3.6. The Liberal Semantic Adjacency Criterion: A Formal Definition

The second additional version is a special variation of the conservative version, which considers SAC for *all* of the temporal relations within the TIRP definition, but enforces it only between symbolic time intervals that represent *different* semantic types.

This new SAC version will allow for the discovery of "Symbolic Gradient" temporal patterns and of "Counting" temporal patterns. This is achieved since it allows patterns that involve successive symbolic time intervals whose symbol includes the same concept, with the same *semantic type* and even with the same *value*, such as $A_3A_2A_1A_1A_1$. In some cases, this expressivity might be useful. This version of SAC is called the *Liberal Semantic Adjacency Criterion*.

**Definition 4.** The *Liberal Semantic Adjacency Criterion* (Liberal SAC or LSAC) holds over a TIRP P = {Ï, Ṙ}, where Ï = $\{I^1, I^2, ..., I^k\}$ iff:

$$\forall i, j, 1 < i < j < k: (I^i.e < I^j.s) \wedge \left(sem\_type(I^i) \neq sem\_type(I^j)\right)$$
$$\rightarrow \nexists t: \left((I^t.s < I^j.s) \wedge (I^t.e > I^i.e)\right)$$
$$\wedge \left(sem\_type(I^i) = sem\_type(I^t) \vee sem\_type(I^j) = sem\_type(I^t)\right)$$

Definition 4 says, in effect, that given the TIRP *P*, there cannot exist in the database, for the same entity, some interval $I^t$ with the same semantic type as $I^i$ or $I^{i+1}$ (where $I^i$ *completely precedes* $I^{i+1}$ in *some* way, directly or indirectly, and $I^i$ has a *different* semantic type from $I^{i+1}$), such that at least *some* of $I^t$ spans at least *some* of the gap between them. This is again essentially the constraint that is exemplified graphically in detail by all of the semantically contradictory instances appearing in Figure 3; but $I^i$, $I^{i+1}$ are not necessarily *successors* in the TIRP's lexicographic ordering (unlike the case of SSAC) and the constraint is enforced only if they are of the *same* semantic type (unlike the case of CSAC).

However, note that, unlike the Sequential SAC version, and certainly unlike the Conservative SAC version, the Liberal SAC constraint *does* allow for some potential semantics that might not be intuitive to domain experts. For instance, the LSAC version not only accepts the TIRP displayed in Figure 5 (and any subset of its components) as legitimate TIRPs, but would also accept the TIRP depicted in Figure 5 as a legitimate TIRP, even if there existed in the same entity's database a "hidden" HIGH Hemoglobin State value for the same entity (e.g., patient), between the two LOW values of the Hemoglobin State abstractions, because the constraint is not enforced between symbolic time intervals that represent the *same* semantic types.

### 3.7. The Computational Implications of Enforcing the SAC

Using the SAC, in addition to enabling the discovery of only patterns whose semantics enforce a stricter interpretation of the data relationships, is potentially also more functional and efficient, i.e., classification and prediction algorithms might benefit from features that represent more "reliable" temporal patterns, and might also enable us to compute them within a briefer time span. Recently, there has been an increasing use of TIRPs as features for classification and prediction [Batal et al., 2009, 2012b; Höppner et al., 2013; Moskovitch and Shahar, 2015b; Patel et al., 2008], in which we would like to examine the potential contribution of SAC.

The SAC is a stricter selection criterion for TIRPs discovery algorithms, and many TIRP candidates might not be generated. Thus, our first hypothesis is that *using the SAC (including the several versions we proposed) to discover TIRPs will generate fewer TIRPs than when not using the SAC for any given minimal vertical support*, which is expected to *lead to a shorter run time for the discovery phase of the TIRPs*.

However, due to the semantic coherence of the discovered TIRPs, i.e., their more uniform meaning, our second hypothesis is that we expect that, *nevertheless, the resultant [smaller] set of TIRPs, discovered using the SAC, will still induce a classifier that has the same or better classification and prediction performance, given the same minimal vertical support threshold for the TIRPs as features*.

### 3.8. Adding the SAC constraint to the KarmaLego algorithm

The SAC is a highly general criterion, equally applicable to any time intervals mining algorithm, as well as sequential mining algorithms. However, we needed to assess it within a concrete framework. We decided to evaluate the SAC within the KarmaLego framework (see Section 3.1). Finally, the algorithm structure's natural modularity, composed of the Karma and Lego steps, greatly facilitated our task of integrating any SAC version.

To implement the SAC version within the KarmaLego algorithm, we first added the basic SAC pruning constraint to the Karma phase. That is, only 2-sized TIRPs that obeyed the basic semi-formal SAC constraint (i.e., of not having any symbolic time interval between them, over which holds a symbol of the same semantic type as either of the two members of the potential 2-sized TIRP) were added to the 2-sized TIRP enumeration tree.

Then, during the Lego phase, given each SAC version to be applied, we decided which pairs of symbolic time intervals needed to be checked against the data when extending the TIRP from size $k$ to size $k+1$:
1. In the case of the SSAC version, we checked the constraint only between the (lexicographically ordered) $k^{th}$ and the new $k^{th}+1$ symbolic time intervals;
2. In the case of the CSAC version, we checked the constraint between the $1^{st}$, $2^{nd}$…, and the (lexicographically ordered) $k^{th}$ and the new $k^{th}+1$ symbolic time intervals;
3. In the case of the LSAC version, we performed a similar procedure to that used to enforce the CSAC constraint, but only for pairs of symbolic time intervals over which hold symbols of different semantic types.

Appendix A contains a short pseudo-code of the SAC implementation with the particular frequent TIRP-discovery algorithm that we had selected, the KarmaLego algorithm. The full implementation details are available elsewhere [Shknevsky, 2014].

## 4. Evaluation

To demonstrate our methods, we decided to use a highly efficient interval-mining algorithm that was recently introduced by the authors, called KarmaLego [Moskovitch and Shahar, 2015a], as our means for discovering TIRPs. However, the semantic enhancements that we introduced into KarmaLego are quite general. We measured the number of discovered TIRPs, the runtime, and the performance of the TIRPs when used as features for several classification and prediction tasks.

We evaluated the runtime of the KarmaLego algorithm and the number of discovered TIRPs with the different SAC versions. Given our informal hypotheses (see Section 3.3), which are based on reasonable arguments, but which need empirical verification, our specific research questions were:

1. Does using SAC indeed reduce the discovery <u>runtime,</u> compared to not using it? Which of the three SAC versions requires the shortest runtime?
2. Does using SAC indeed reduce the <u>number of discovered TIRPs</u>, compared to not using it? Which of the three SAC versions results in the smallest number of TIRPs?
3. Does using SAC maintain the <u>classification and prediction performance</u>, compared to not using it? Which of the SAC versions is best for classification and prediction?

This evaluation was performed across different state abstraction or discretization methods (KB, EWD, SAX, and TD4C-KL, which uses the Kullback–Leibler distance measure as explained in Section 2.1), each with 3 bins, different temporal relation sets (the three abstract temporal relations mentioned in Section 2.2, and the full set of Allen's seven temporal relations), and various minimal vertical support thresholds to measure the number of discovered TIRPs. In addition, for the purpose of the classification and prediction tasks, we evaluated different TIRP feature-representation methods (Binary, Horizontal Support, and Mean Duration) (as discussed in Sections 2.3 and 3.1) and different classification algorithms (Random Forest, Naïve Bayes, Support Vector Machines [SVM], and Logistic Regression). We expand on our classification-performance evaluation methods in Section 4.2.

## 4.1. The Data Sets

To evaluate the effect of using the SAC versions on the results of the TIRP discovery process and on the eventual performance of the models induced for classification and prediction, we used three clinical datasets.

The data sets included: (1) an oncology dataset from the Rush Medical Center, Chicago, USA, including patients who had undergone either allogeneic or autologous bone-marrow transplantation; (2) a hepatitis data set describing patients who had either Hepatitis B or C, from a KDD conference challenge [Ho and Nguyen, 2003], which is *publicly available* [Berka et al., 2002]; and (3) a diabetes dataset from our local academic medical center [Gordon, 2012; Klimov et al., 2015], including patients who had been followed (albeit sporadically) for at least five years, focusing on the future outcome of the level of albuminuria (protein in the urine, a measure of renal dysfunction) in the fifth year. (Only the Hepatitis data set is publicly available.)

Table 1 describes the characteristics of the three data sets used throughout all of the evaluations: the total number of data points, the number of patients, the number of concepts, or the number of all semantic types (e.g., Hemoglobin State in a particular context, as explained in Section 3.3), and the average number of data points per patient.

Table 1 - Descriptive statistics of the three datasets.

| Dataset   | Data Points | Patients | Concepts | Mean Data Points Per Patient |
|-----------|-------------|----------|----------|------------------------------|
| Oncology  | 76,468      | 207      | 12       | 369                          |
| Hepatitis | 368,216     | 499      | 10       | 738                          |
| Diabetes  | 165,199     | 5178     | 4        | 32                           |

The task in the case of the oncology dataset was to classify patients who underwent bone-marrow transplantation into autologous bone-marrow transplantation versus an allogeneic bone-marrow transplantation; the task in the case of the hepatitis dataset was to classify the patients into Hepatitis B patients versus Hepatitis C patients; and the task in the case of the diabetes dataset was the prediction, within a variable period of up to 5 years, of the state abstraction of the albuminuria-value concept (a measure of the amount of protein secreted in the urine), and specifically, whether the patient will have a normal albuminuria-level (denoting normal renal function) versus a micro-albuminuria or macro-albuminuria albuminuria-level (indicating renal deterioration).

The full description of the three data sets, the definitions used within each domain, in the case of the knowledge-based temporal state abstraction method, and additional details about the tasks within each domain, appear in Appendix B.

## 4.2. The Experimental design and the Evaluation Measures

We based our evaluation on the KarmaLegoSification framework [Moskovitch and Shahar, 2015b]; The input time-stamped raw data were interpolated and abstracted into a set of symbolic time intervals, using either knowledge-based or automatic temporal abstraction methods. All of the frequent TIRPs that can be discovered were discovered from the symbolic time intervals output, with or without using any SAC version. In either case, we examined the effect of using either the abstract three temporal relations or the full seven temporal relations. The TIRPs were then used as features for the induction of classification and prediction models, by representing each TIRP using either a simple binary representation of the TIRPs, the mean horizontal support, or the mean duration of the TIRP within the entities.

For the evaluation of research questions 1 and 2 we performed a series of experiments recording the runtime in seconds and the number of discovered TIRPs using the KarmaLego algorithm with the three SAC versions, as well as without using any criterion at all. Note that we could use any other temporal data mining algorithm; we used KarmaLego because it is faster than several other approaches and it is complete [Moskovitch and Shahar, 2015a]. The runtime and number of TIRPs were evaluated on the different temporal abstraction methods, different sets of relations, and various minimal vertical support thresholds.

Because these experiments measure runtime, each combination was executed separately and thus was isolated from other processes that might have influenced the CPU behavior. We used an AMD Opteron™ Processor 6128 2.00 GHz Machine with 32.00 GB RAM and Windows Server 2008 R2 Datacenter.

To answer research question 3, we evaluated the classification and prediction performance of the SAC using the *Area Under the Curve* (AUC). We compared the mean AUC with two statistical analysis methods: a one-way ANOVA and the post hoc Scheffé method, using IBM SPSS Statistics 20. The one-way ANOVA was applied to the general parameters, such as determining whether the different SAC versions performed differently and the Scheffé method was applied as post hoc examination to test differences within the SAC versions. Comparisons that were found to be significantly different ($\alpha = 0.05$) are reported.

Since mining TIRPs may result with different sets for each group of patients [Moskovitch and Shahar, 2015b] we used a rigorous evaluation setup, including three-fold mining and ten-fold cross-validation classification. Thus, the data were split into three folds; then TIRPs were discovered from one fold and were detected in the other two folds, which were used for the classification experiment. This was repeated three times for the three-folds mining. We used four highly different types of induction algorithms: *Random Forest*, the best known application of the decision trees family (randomizing both the features and the data) [Breiman, 2001], the classic pure probabilistic reasoning algorithm – *Naïve Bayes* [John and Langley, 1995], SVM – a very different family that uses a special type of linear optimization [Keerthi et al., 2001], and of course from the Linear classifiers, *Logistic Regression* [Landwehr, 2005], which is often used as the baseline statistical approach against which other methods are compared.

The KarmaLego method was implemented based on the original Moskovitch and Shahar study [Moskovitch and Shahar, 2015a]. We used the SAX algorithm, which we implemented based on Lin et al.'s description [Lin et al., 2003], and the TD4C-KL method, which was implemented based on Moskovitch and Shahar [Moskovitch and Shahar, 2015c], using the Kullback–Leibler divergence as the measure for deciding which value cut-off leads to the best separation between the outcome classes. We used the classification algorithm implementations available in WEKA 3.7.1 [Frank et al., 2005].

## 5. Results

### 5.1. The SAC runtime and number of discovered TIRPs

For each data set we ran the experiments with various minimal vertical support thresholds (for reasonable runtime and memory usage).

## 5.1.1 The Oncology dataset

Figure 6 presents the runtime in seconds of the KarmaLego algorithm using Allen's seven temporal relations, and Figure 7 using the abstract three relations.

Figure 6 and Figure 7 show that all of the SAC versions result in a faster runtime. Using SAC (and especially CSAC) allowed us to compute all of the TIRPs passing the minimal vertical support threshold in a time that was almost an order of magnitude shorter than the time needed without using the SAC version.

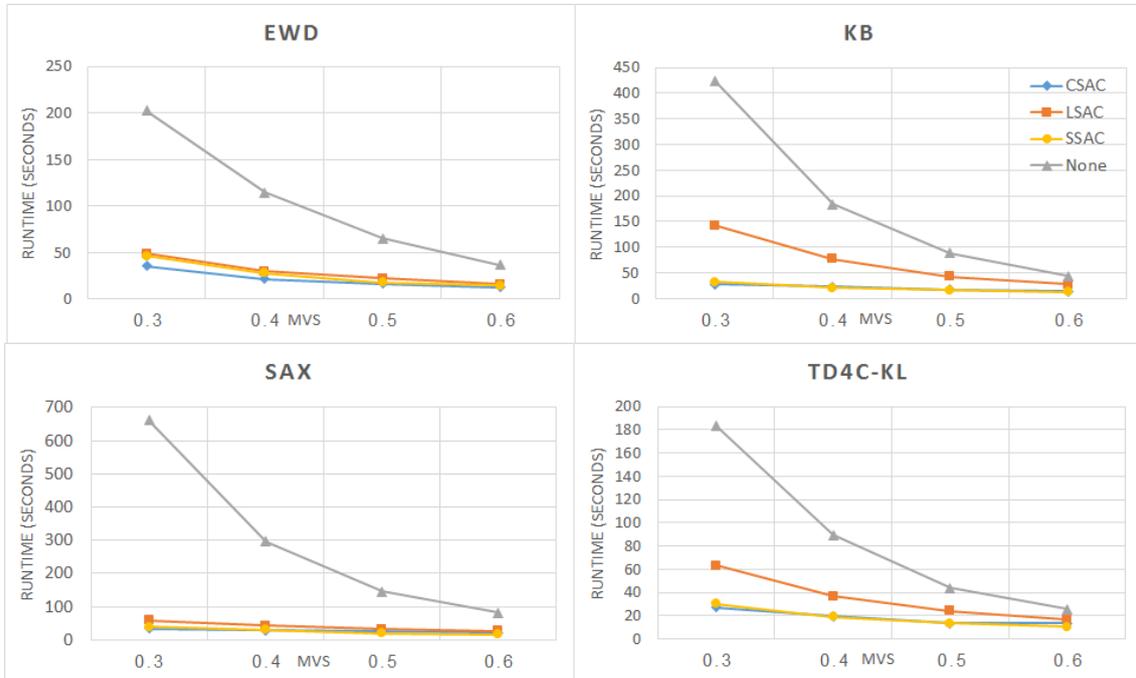

**Figure 6** - The runtime of the KarmaLego algorithm on data mined using seven temporal relations in the oncology dataset. Each graph represents one temporal abstraction method, and displays all of the SAC versions (if any) used (the legend appears on upper right)

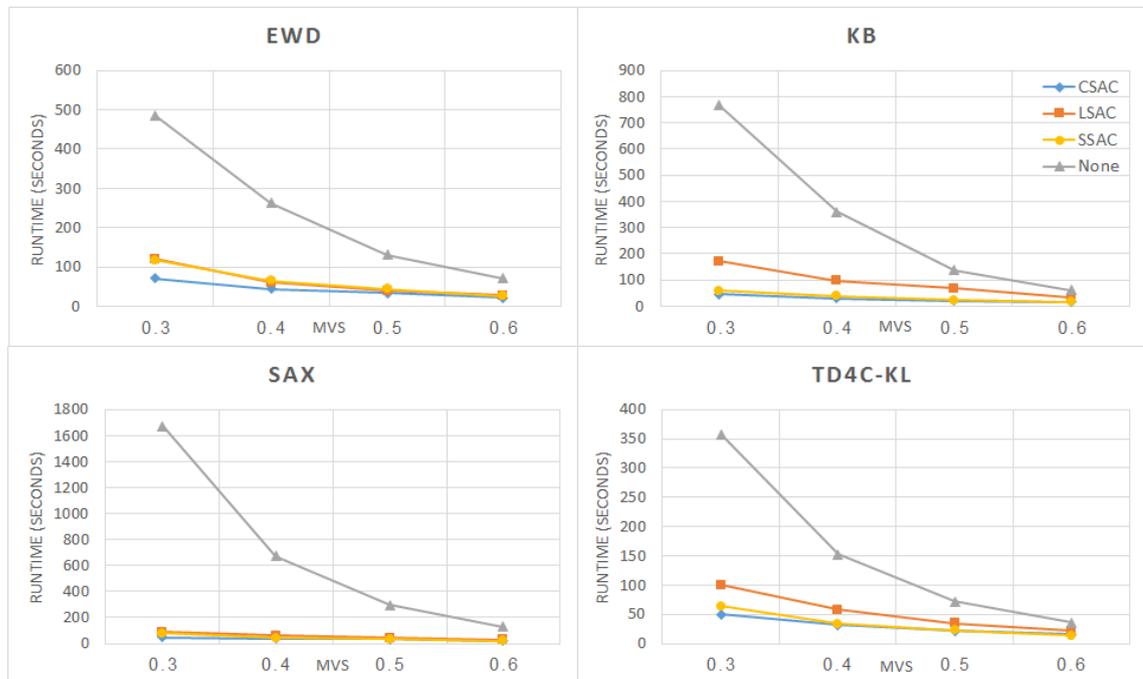

**Figure 7** - The runtime of the KarmaLego algorithm on data mined using three temporal relations in the oncology dataset. Each graph represents one temporal abstraction method, and displays all of the SAC versions (if any) used (the legend appears on upper right)

Figure 8 presents the number of discovered TIRPs when the data was abstracted using Allen's seven temporal relations, and Figure 9 when the data abstracted using the abstract three relations. The same trends, as for runtime, hold for the number of discovered TIRPs.

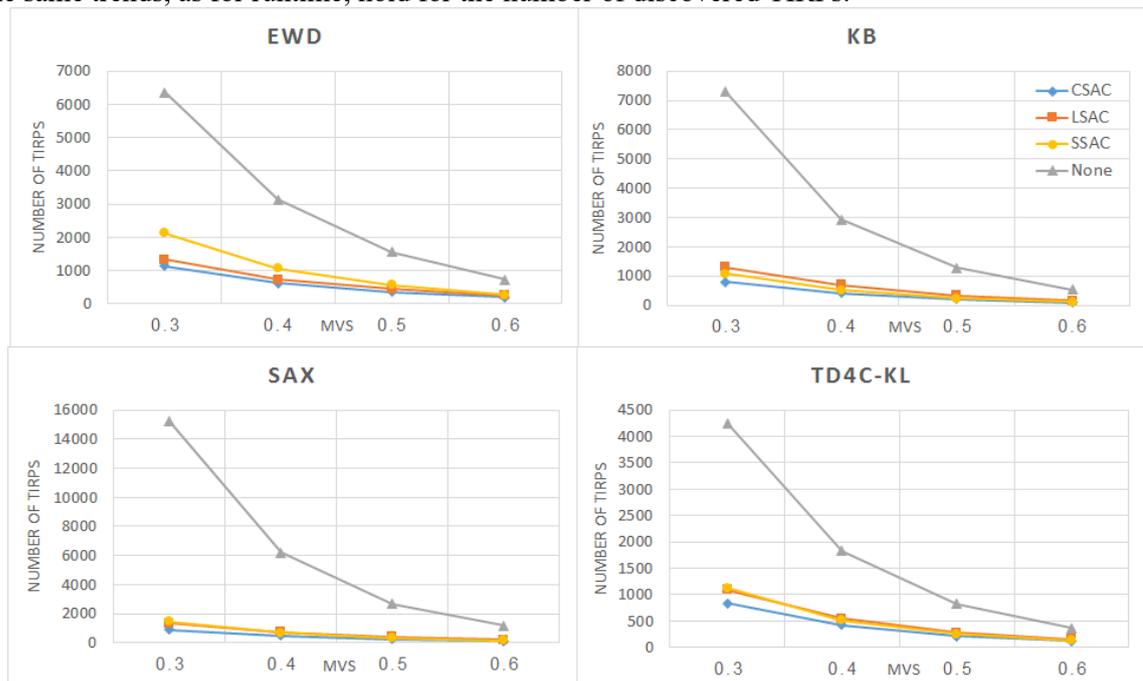

**Figure 8** - The number of discovered TIRPs using seven temporal relations in the oncology dataset. Each graph represents one temporal abstraction method, and displays all of the SAC versions (if any) used (the legend appears on upper right)

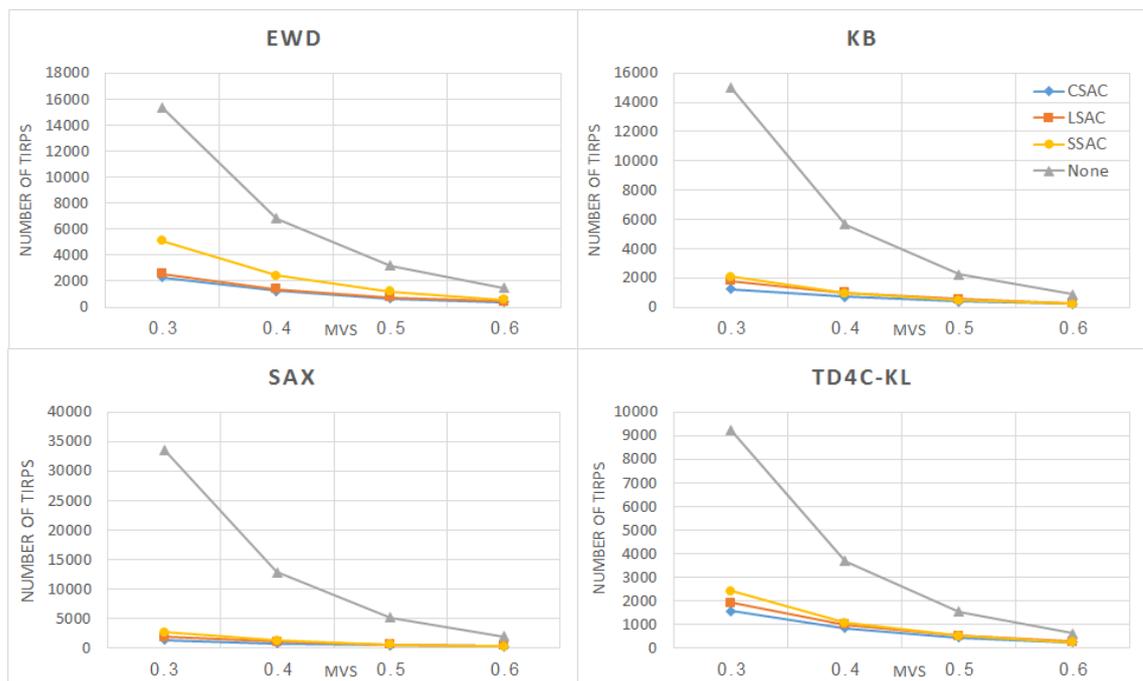

**Figure 9** - The number of discovered TIRPs using three temporal relations in the oncology dataset. Each graph represents one temporal abstraction method, and displays all of the SAC versions (if any) used (the legend appears on upper right)

### 5.1.2 The Hepatitis dataset

Figure 10 presents the runtime in seconds of the KarmaLego algorithm using Allen's seven temporal relations, and Figure 11 using the abstract three relations. Using LSAC and CSAC allowed us to compute all of the TIRPs passing the minimal vertical support threshold in a time that was almost an order of magnitude shorter than the time needed using SSAC or without using any SAC. The most restrictive CSAC version is the fastest.

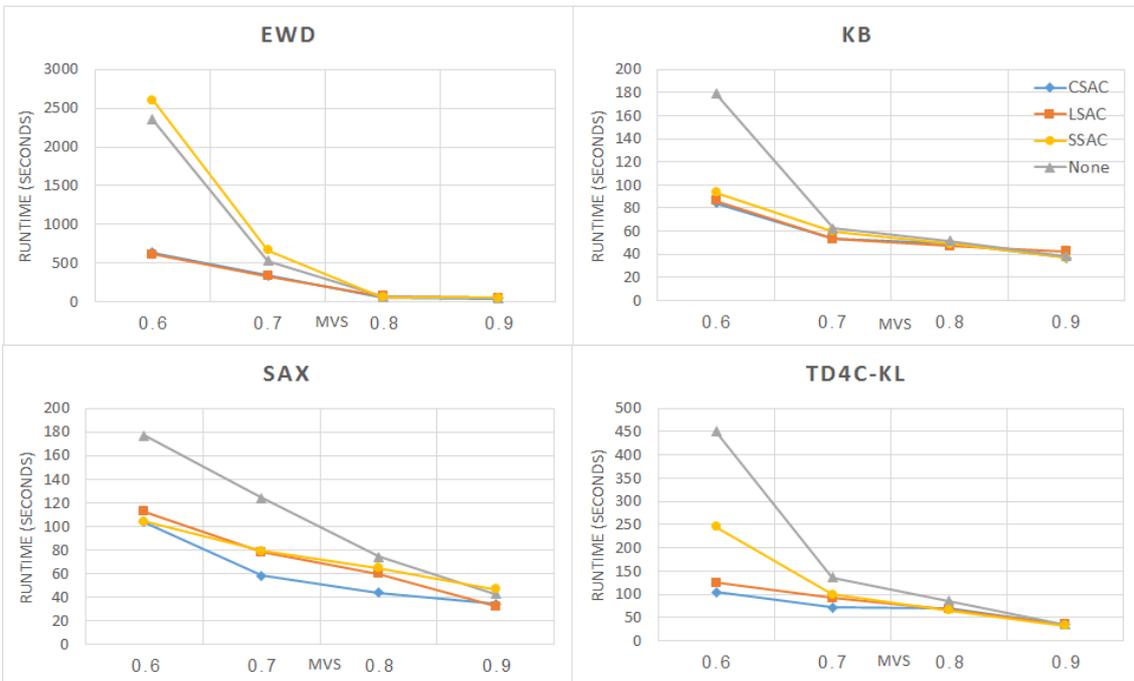

**Figure 10** - The runtime of the KarmaLego algorithm on data mined using seven temporal relations in the hepatitis dataset. Each graph represents one temporal abstraction method, and displays all of the SAC versions (if any) used (the legend appears on upper right)

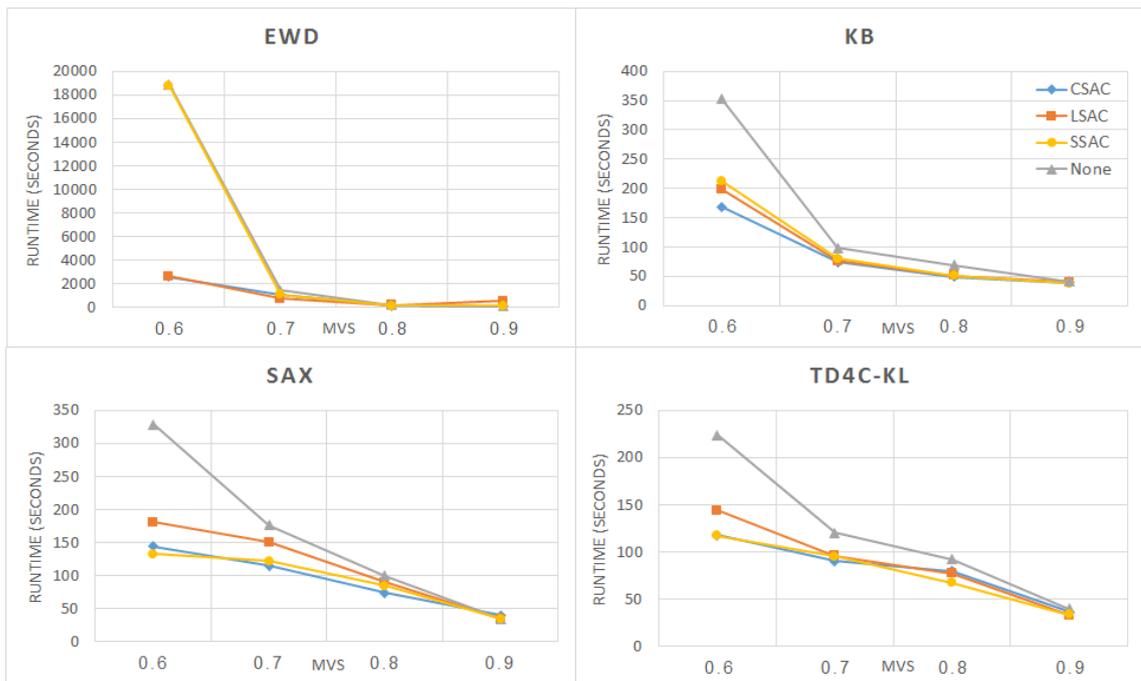

**Figure 11** - The runtime of the KarmaLego algorithm on data mined using three temporal relations in the hepatitis dataset. Each graph represents one temporal abstraction method, and displays all of the SAC versions (if any) used (the legend appears on upper right)

Figure 12 presents the number of discovered TIRPs when the data were mined using Allen's seven temporal relations, and Figure 13 when the data were mined using the abstract three relations. Here too, the same trends are seen as in the runtime results: Using the SAC versions usually results in the discovery of a significantly smaller number of TIRPs and within a shorter runtime.

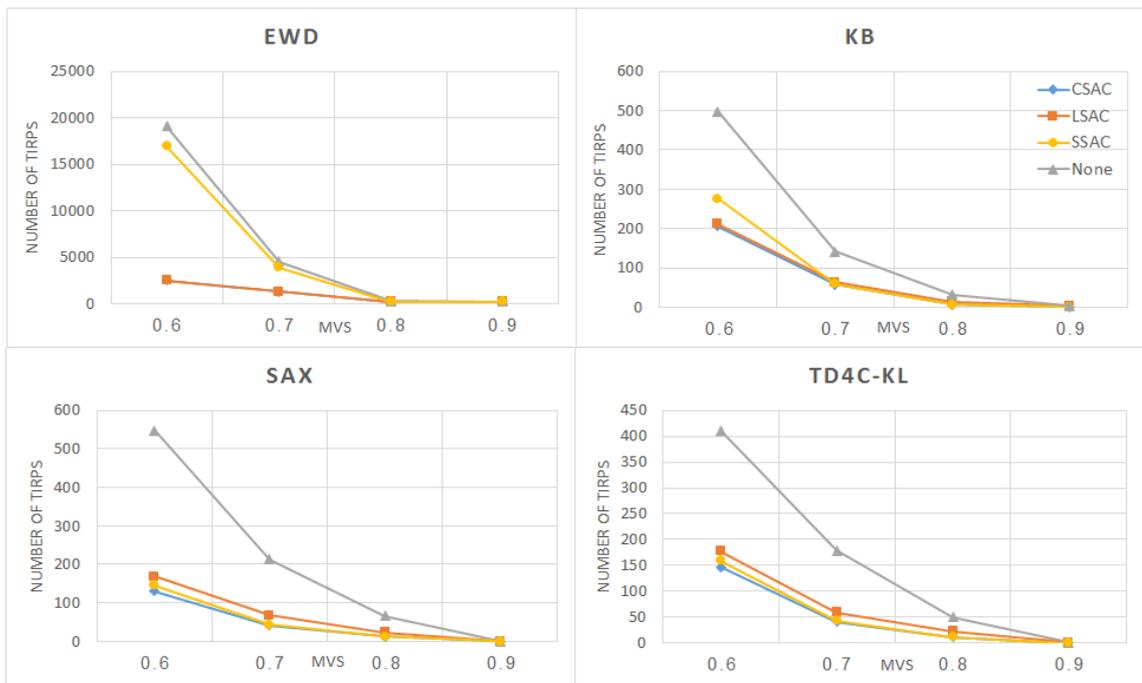

**Figure 12** - The number of discovered TIRPs using seven temporal relations in the hepatitis dataset. Each graph represents one temporal abstraction method, and displays for each method all of the SAC versions (if any) used (the legend appears on upper right)

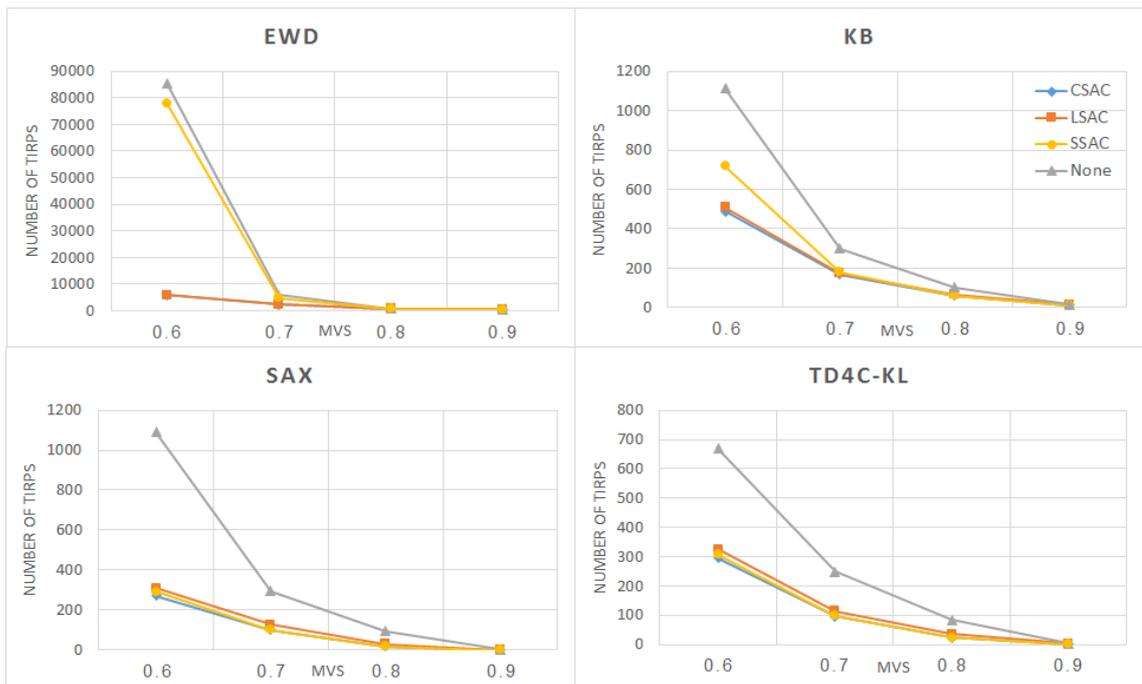

**Figure 13** - The number of discovered TIRPs using three temporal relations in the hepatitis dataset. Each graph represents one temporal abstraction method, and displays all of the SAC versions (if any) used (the legend appears on upper right)

### 5.1.3 The Diabetes dataset

Figure 14 presents the runtime in seconds of the KarmaLego algorithm using Allen's seven temporal relations, and Figure 15 when using the abstract three relations. From both figures we can see that using the SSAC and CSAC versions results in a faster extraction of the TIRPs. The most restrictive version, CSAC, was also the fastest, as would be expected. Using SSAC and CSAC allowed us to compute all of the TIRPs passing the minimal vertical support threshold within a much shorter time.

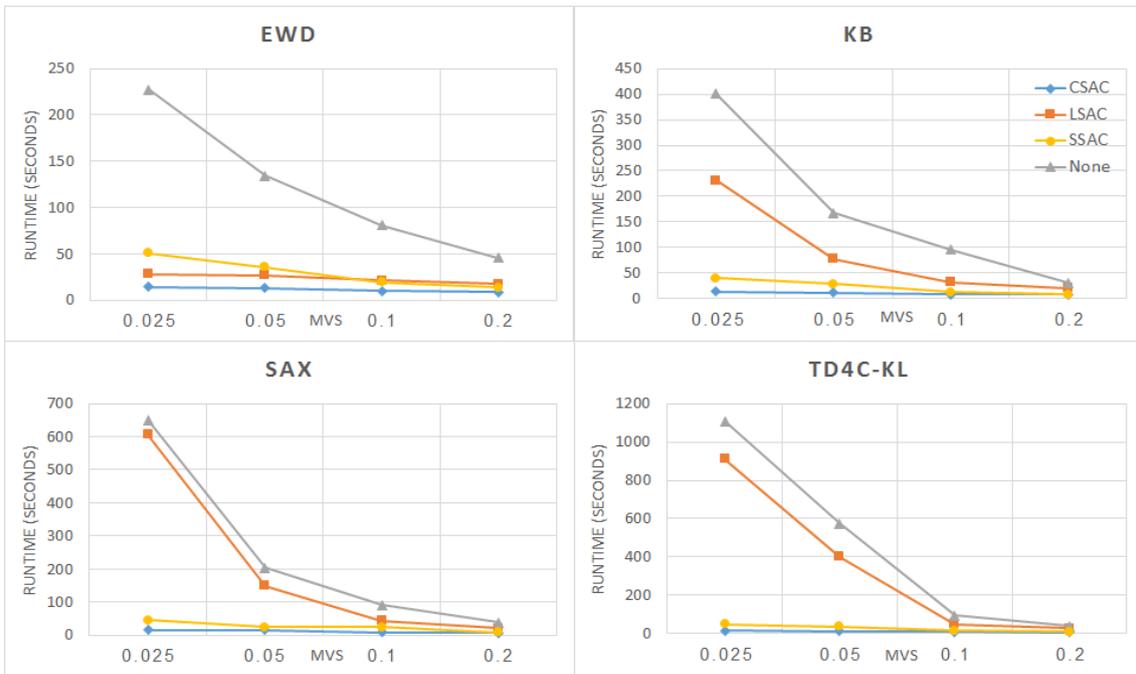

**Figure 14** - The runtime of the KarmaLego algorithm on data mined using seven temporal relations in the diabetes dataset. Each graph represents one temporal abstraction method, and displays all of the SAC versions (if any) used (the legend appears on upper right)

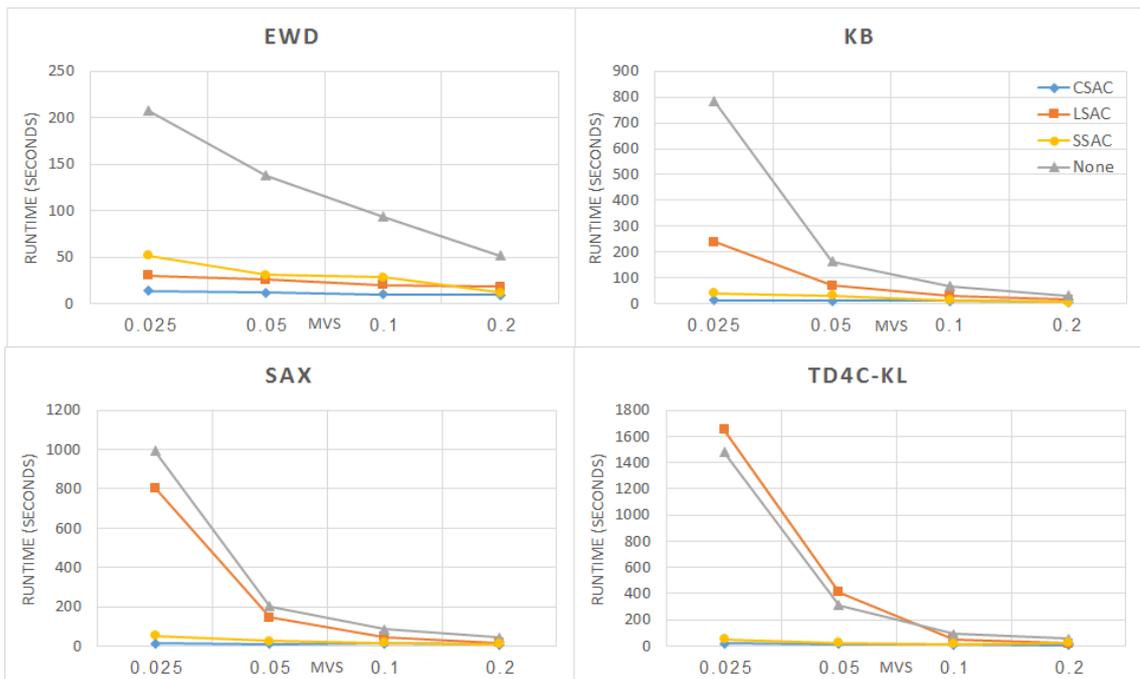

**Figure 15** - The runtime of the KarmaLego algorithm on data mined using three temporal relations in the diabetes dataset. Each graph represents one temporal abstraction method, and displays all of the SAC versions (if any) used (the legend appears on upper right)

Figure 16 presents the number of discovered TIRPs of the KarmaLego algorithm when the data were mined using Allen's seven temporal relations, and Figure 17 when the data were mined using the abstract three relations. All SAC versions resulted in fewer TIRPs when compared to the number of non-SAC-obeying TIRPs.

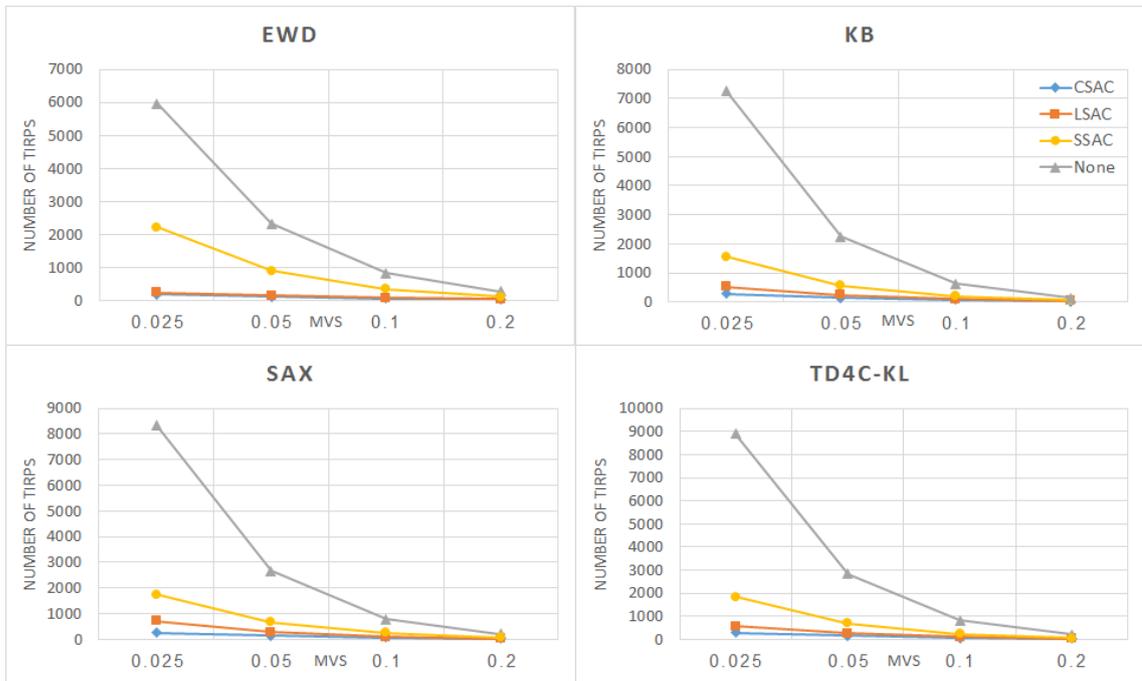

**Figure 16** - The number of discovered TIRPs, using seven temporal relations in the diabetes dataset. Each graph represents one temporal abstraction method, and displays all of the SAC versions (if any) used (the legend appears on upper right)

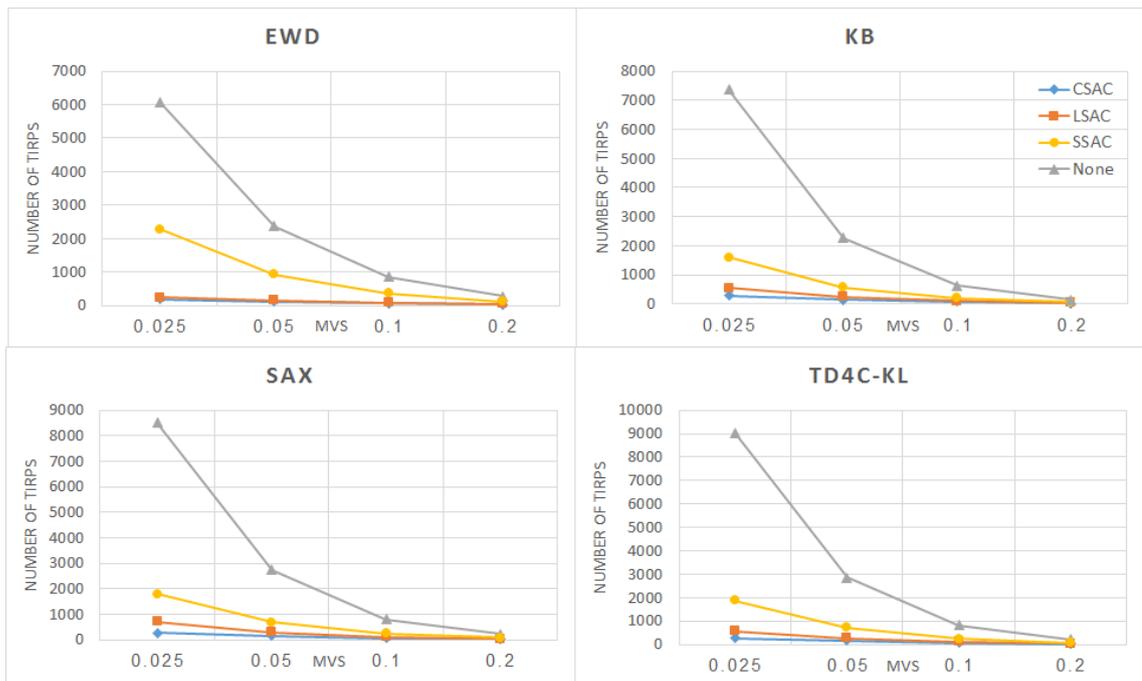

**Figure 17** - The number of discovered TIRPs, using three temporal relations in the diabetes dataset. Each graph represents one temporal abstraction method, and displays of the SAC versions (if any) used (the legend appears on upper right)

We saw the best results in the diabetes dataset, which is also the largest one. When running the experiment with 0.025 minimal vertical support, we discovered 7689 patterns (in about 731 seconds) when *not* using SAC, and only 253 patterns (in only about 15 seconds) when using CSAC. Thus we got up to a 97% decrease in the number of discovered patterns in up to 98% less time. Overall, discovering SAC-obeying TIRPs is faster and the number of discovered TIRPs is much smaller when using the SAC. Moreover, the most restrictive CSAC version resulted with fewer TIRPs and the fastest runtime.

## 5.2. Classification and prediction performance using the SAC

For each dataset we calculated temporal abstractions based on the KB, EWD, SAX, and TD4C-KL temporal abstraction methods. We then discovered frequent TIRPs that are composed of the temporal abstractions and temporal relations among them, using the KarmaLego algorithm, with or without the SAC enhancement. To generate TIRPs, we examined both the use of Allen's full seven temporal relations as well as the use of only the three abstract temporal relations. The TIRPs were used as features to train a classifier using the various induction methods. Note that to produce the TIRP features, in each data set we used a different minimal vertical support, such that it produced features that characterize at least half of the patients, or at least produced a reasonable number of features (tens to hundreds of features).

We then trained a classifier using each of the four classifier-induction methods we had chosen (*Random Forest*, *Naïve Bayes*, *SVM*, and *Logistic Regression*), and evaluated the performance of the resultant classifiers using the methodology explained in Section 4.2.

Figure 18 displays the mean result of using the four resultant classifiers in the three domains, when using any of the three SAC versions during the TIRP discovery process, compared to the classification results when the TIRP discovery process did not use any SAC version (the results are averaged over all the SAC versions, representation methods, temporal abstractions, and temporal relations variations). As can be seen, using the greatly reduced set of discovered TIRPs achieved at least the same classification (or prediction) performance in all of the tested configurations as when using the original, full set of TIRPs (without any SAC-based pruning). The classification performance, in each version of the experiment, was evaluated using the Binary (B), Horizontal Support (HS), or Mean Duration (MeanD) TIRP representation methods.

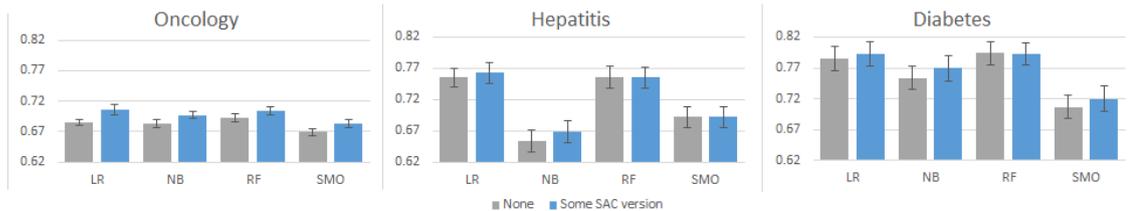

**Figure 18** - The mean AUC (Y-axis) of using the four classifier-induction methods in all three domains, when using any of the three SAC versions during the TIRP discovery process, compared to not using any SAC version during that process. RF=Random Forest; NB=Naïve Bayes; SMO=Support Vector Machine; LR=Logistic Regression

To present the results in more detail, we focus in the rest of this section only on the results of the *Random Forest* algorithm, because (a) very similar results were achieved for all of the classifier-induction methods, with respect to the effectiveness of either using, or not using, the SAC enhancement, and we wanted to avoid a tedious repetition; and (b) its overall classification performance in the baseline case, when *not* using the SAC enhancement, was slightly better, in a consistent fashion, than that of the other induction methods.

### 5.2.1 The Oncology dataset

For the oncology dataset we used a minimal vertical support of 0.5. All SAC versions performed the same, regarding classification accuracy, as when not using a SAC, in spite of using a much smaller number of TIRPs when using the SAC. We can see from our results of the empirical evaluation that using the SAC led to a slightly better performance, no matter which TIRP representation (see Figure 19) or abstraction method (see Figure 20) was used, although the differences are not significant. SAC also performed slightly better when using either three or seven temporal relations.

Note that each point in the figures represents the average AUC of multiple runs (see Evaluation Methods). For example, in Figure 19, each point represents the mean of 240 different experimental runs (3 pattern extractions from 1/3 of the data each time × 10 folds × 4 abstraction methods × 2 sets of temporal relations). Figure 20 points represent 180 different experimental runs (3 pattern extractions from 1/3 of the data each time × 10 folds × 3 feature representation methods × 2 sets of temporal relations).

The number of the TIRPs discovered without the use of the SAC was meaningfully larger. However, it did not result in a superior classification performance, compared to the use of the reduced sets of TIRPs that resulted when using the SAC. The full set of TIRPs did not have additional classification or prediction power; rather, it even slightly reduced the performance. (See the Discussion section for several possible implications.)

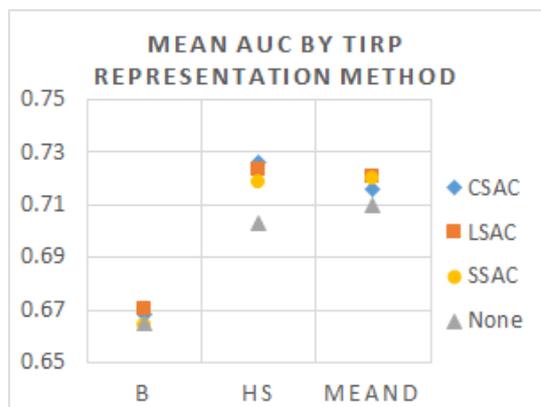
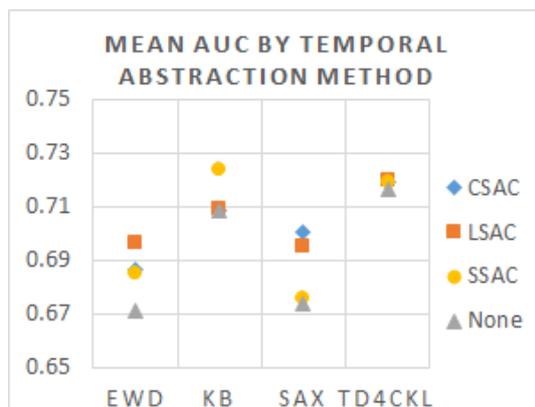

**Figure 19** - The classification performance results when using the different SAC versions in the oncology dataset partitioned by TIRP representation methods

**Figure 20** - The classification performance results when using the different SAC versions in the oncology dataset partitioned by the temporal abstraction methods

### 5.2.2 The Hepatitis dataset

For the hepatitis dataset we used a minimal vertical support of 0.7. With and without all SAC versions performed the same (see Figure 21 and Figure 22), in spite of the use of a much smaller set of discovered TIRPs. No significant difference between using three or seven temporal relations with respect to classification performance was found.

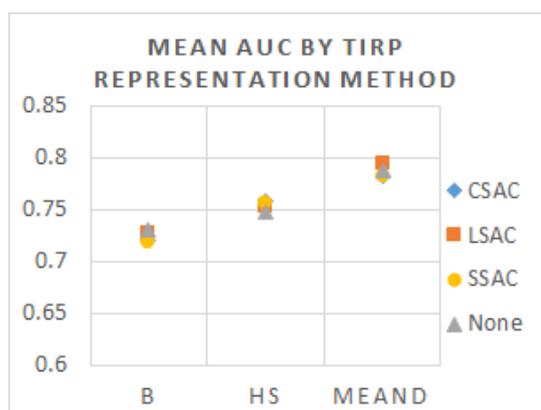
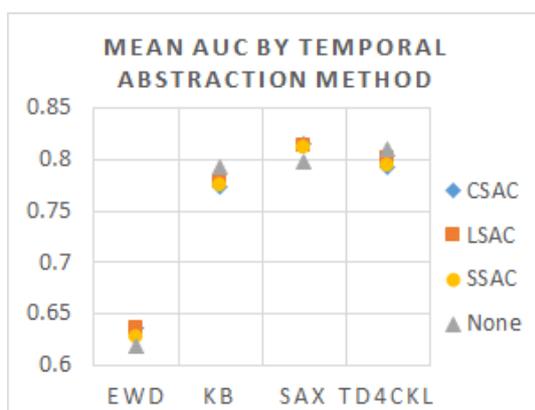

**Figure 21** - The classification performance results when using the different SAC versions in the hepatitis dataset partitioned by TIRP representation methods

**Figure 22** - The classification performance results when using the different SAC versions in the hepatitis dataset partitioned by the temporal abstraction methods

Although the dataset is dense, and there are multiple instances of the same TIRP for each patient (as opposed to the sparser oncology dataset), using the reduced set of TIRPS as features led to a classification performance that was as good as that of using the large set of TIRPs discovered without using the SAC. Thus, using the full set of TIRPs was not superior over the reduced set of TIRPs discovered using the SAC.

### 5.2.3 The Diabetes dataset

For the diabetes dataset, we used a minimal vertical support of 0.1. Using the features discovered by using all three SAC versions, and the original TIRP discovery process without using SAC, led to a similar level of prediction performance (see Figure 23 and Figure 24), in spite of the use of a much smaller set of discovered TIRPs. There was no significant difference in performance when using three versus seven temporal relations.

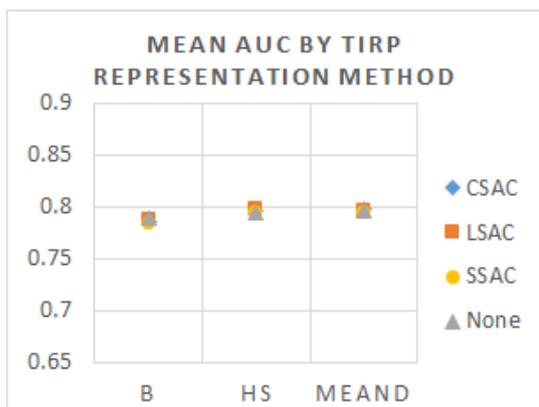 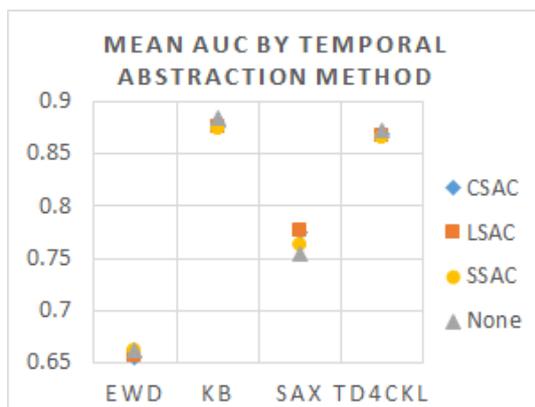

**Figure 23** - The prediction performance results when using the different SAC versions in the diabetes dataset partitioned by TIRP representation methods

**Figure 24** - The prediction performance results when using the different SAC versions in the diabetes dataset partitioned by the temporal abstraction methods

Using the full set of TIRPs did not seem to have any additional benefit regarding prediction performance; it was sufficient to use the reduced set of TIRPs discovered using the SAC. This data set is sparser in comparison to the two others; thus, there are fewer SAC-obeying TIRPs, but the performance stays the same. In this domain, the maximal gap for the *before* relation was the largest, and thus, a larger number of potential semantic contradictions may have been avoided, compared to not using any semantic considerations; thus, using the SAC led to a good performance regardless of the small number of discovered TIRPs, which was an order of magnitude less than when not using any semantic considerations. From a practical point of view, it seems that CSAC produced the smallest set of TIRPs while maintaining the classification and prediction performance, and even less when using the seven temporal relations. In parallel, Mean Duration TIRP representation method and the Knowledge-based discretization method were the best for classification and prediction tasks. In summary, the reduced set of discovered TIRPs, with a much faster discovery runtime, using the various versions of the SAC, maintained the same classification (or prediction) performance in all of the tested configurations. These results complement the result of discovering a much smaller number of TIRPs when using the SAC. Note also the clear trend towards higher performance when using the various SAC versions, in spite of the much reduced feature set.

## 6. Discussion and Conclusions

In this study we have defined, formalized, and assessed in detail the computational significance of the *Semantic Adjacency Criterion*, a TDM filtering principle that in the past we had only briefly and informally introduced.

Note that *no new temporal patterns were discovered*, of course, during the filtering-out process using the SAC principle, but the number of temporal patterns found was greatly reduced by the various SAC pruning versions that we had used. Three versions of the SAC criterion were defined and analyzed in detail: The *Sequential Semantic Adjacency Criterion* (SSAC) [Shknevsky et al., 2014] which enforces the constraint only over pairs of temporally successive symbolic time intervals within the TIRP's definition; the *Conservative Semantic Adjacency Criterion* (CSAC), which enforces the constraint over *every* pair of symbolic time intervals within the TIRP's definition (recall that in a TIRP, a temporal relation is defined in an unambiguous fashion between *each* pair of symbolic time intervals); and the *Liberal Semantic Adjacency Criterion* (LSAC), a variation of CSAC, which enforces the constraint only over pairs of symbolic time intervals that have *different* semantic types.

It is important to note that each type of SAC can serve a different purpose and has a different expressivity. For instance, the LSAC version implicitly enables a *counting* of symbolic time intervals of the same type (e.g., patterns such as A1A1A1, denoting, for example, three overall administrations of the same range of the dose of the medication, although different ranges of doses of the medication might have been administered between them), while constraining intervals of different semantic types. Another example is the SSAC version, which enables a more restricted version of counting, but, like the LSAC version, enables the discovery of a TIRP that implicitly contains a "Symbolic Gradient" temporal pattern (e.g., patterns such as A1A2A3B, denoting, for example, a Low level of the medication-dose, followed by a Medium level and then a High level, followed by some side effect). The CSAC version seems the most useful to maintain compactness in the number of discovered TIRPs, while preserving the strictest semantics of the TIRPs, although it prevents the discovery of repetitions of symbolic time intervals within the same TIRP, or the discovery of Symbolic Gradients.

Note also that as the minimal vertical support increases, and thus the number of potentially discoverable patterns decreases, the returns for using the SAC principle are diminishing; however, this is exactly the phenomenon that might enable researchers and physicians to extract useful patterns, using the SAC principle, from bigger, and even very big, datasets.

We evaluated the classification and prediction performance of the features discovered using the three SAC versions on three different medical domains: oncology, infectious hepatitis, and type II diabetes. *Note that we did not choose any simulated or artificial data set,* since the main point of the evaluation was to test the SAC principle within *real* clinical domains that incorporate *real semantics*, and in particular, potential *causality*. We believe that the true value of the SAC principle can only be apparent within *real-world* data, since it is precisely the lack of coherence of most temporal patterns that is being filtered out by that principle.

To further bolster the assessment process and its conclusions, we performed the evaluation using classification algorithms from *four different classifier-induction families*: Random Forest, Naïve Bayes, SVM, and Logistic Regression.

It is important to emphasize that *the main objective of this study was to demonstrate the possibility of reducing the number of pattern features that need to be discovered in a large time-oriented data set by at least an order of magnitude (and enhance their intuitive meaning to a domain expert, due to their increased transparency), without losing any classification performance.* Our goal here was *not* to *enhance* any classification performance (although, as we shall see, that might be a future outcome of the current study).

The different SAC versions behaved slightly differently in each domain and for each classifier version. But overall, using all of them required much less *time*, up to 98% less than when not using any SAC version, depending on the minimal vertical support specified, to discover all of their respectively relevant TIRPs, compared to not using any SAC at all. Using the various SAC versions also resulted in a significantly reduced *number* of TIRPs, up to 97% less, depending on the minimal vertical support threshold. This reduced set of TIRPs, however, did not lead to any reduced performance in any of the three medical domains, i.e., the resulting classifiers performed as well when using this reduced set of features as when using the full original set of TIRPs discovered in the standard KarmaLego methodology.

We infer that, at least in the medical domains in which we assessed our methodology, SAC-obeying TIRPs seem to contain most of the information important for classification and prediction.

Most of the data sets we used were relatively small (at least compared to current big-data sets, although the data sets we experimented with contained about 70,000, 160,000, and 360,000 data points). However, the clear trend noted above towards a *higher* performance when using the *reduced* set of TIRPs, filtered using various SAC versions, in spite of the much smaller feature set, suggests that repeating our studies with much larger data sets, might, in fact, not only show that the much smaller set of TIRP-based features is sufficient, but might even demonstrate a significant *improvement* in the classification performance; future studies might elucidate that aspect.

However, in any case, reduction of the number of temporal pattern features to be discovered in a big data set has significant computational implications. On a similar note, it is interesting to consider that Fradkin and Mörchen's conclusions in a different TDM study [Fradkin and Mörchen, 2015] were that the main advantage of their proposed sequential mining algorithm, BIDE-DC, lies in generating a *smaller number of patterns*, while preserving the same classification performance.

As we noted in the Introduction, we have previously shown that frequent TIRPs can be consistently discovered, and in similar proportions, in different subsets of the same data set, within three different medical domains, thus increasing their value for potential patient trajectory clustering, classification, and prediction tasks [Shknevsky et al., 2017]. (In fact, we used the same three domains mentioned in the current study.) The study has also shown that consistent discovery can be increased by increasing the minimal-support threshold for frequency, and, interestingly, by using the SAC principle to prune in each subset the patterns that are candidates for discovery.

Note also that the SAC principle is quite general, and is not specific to the KarmaLego algorithm on which we demonstrated it or even to the family of multivariate interval-based TDM algorithms that KarmaLego is a part of. For example, sequential mining algorithms such as SPADE [Zaki, 2001] start with a set of time-stamped events, each containing several items, and discover qualitative associations that involve the Before temporal relation. Using SPADE to generate a set of temporal sequences that will be used as classification features might well benefit similarly through the addition of semantic considerations similar to the SAC variations, while enhancing its semantic transparency to domain experts.

## 6.1. Limitations

We did not measure the absolute runtime of the classification and prediction phases, but obviously, representing a larger number of features, especially when using various functional methods (e.g., computing the mean duration of each TIRP) requires more time. That might be an additional advantage of using the SAC principle, which we did not assess. Other factors that might also require more time are selecting and using various feature selection algorithms, and inducing a classifier from a larger set of features.

Note that a trivial case for semantic equivalence is the one in which all concepts are different (e.g., different events, each with its own symbol); semantic type equivalence between two symbolic intervals will then consist of having the same symbol hold over both intervals.

Our main intent in the current study, however, was to explore the non-trivial case in which most concepts might have *more than one value*, or, at least, in which there is some domain knowledge that assigns types to the various concepts. However, exploring the potential implications of the SAC principle to the simple case in which all symbols are different and no domain knowledge exists can certainly be explored in a future study.

We did not assess the actual transparency of the SAC-obeying TIRPs, as opposed to non-SAC-obeying TIRPs, in the eyes of medical domain experts in the three domains. That was not an objective of the current study, but it might be interesting to assess that explicitly in future studies.

The use of the four different abstraction and discretization methods led *qualitatively* to the same results, with respect to the number of TIRPs discovered, the time needed to discover them, and the performance of the TIRPs as features for classification and prediction purposes, in all three domains, using four different classification algorithms. Nevertheless, when using the EWD abstraction method, we noted in the specific case of the hepatitis dataset that the SSAC's runtime (and the number of discovered TIRPs) was close to the runtime achieved without the use of any SAC (see Figure 10 and Figure 11). SSAC is a sequential version of SAC; thus, the most reasonable explanation for this phenomenon is probably that the hepatitis data were not sequential, and most of the concepts appeared at the same time. Still, the use of SSAC did not significantly reduce the performance compared to not using any semantic criterion.

Not all of the SAC versions performed equally well in the case of the diabetes dataset (see Figure 14 and Figure 15). Using LSAC, which is the liberal version of SAC, meaning that it restricts the criterion to hold only over pairs of semantically *different* concepts, led to a worse performance when compared to the other SAC versions. The reason might be that the diabetes dataset includes a small number of concepts measured repeatedly over a long time, and is pretty sparse; but there are several laboratory tests that are very common and, in the case of the liberal version of the SAC, relations among pairs of intervals of the *same* concept were considered, just as in the case of not using any semantic criterion; the result is a runtime that is pretty close to that of not using *any* semantic considerations, at least in some of the configurations.

The last two examples, i.e., the exceptions in our results of the empirical evaluation, demonstrate that one must learn the data and select the most appropriate SAC version, as well as the other parameters, e.g., discretization and representation methods. However, overall, the CSAC version performed best, no matter which configuration was chosen.

## 6.2. Conclusions

We defined and formalized in detail a new *Semantic Adjacency Criterion* for TIRP discovery, which increases the transparency of the discovered TIRPs for domain experts, and which can exploit even very basic domain knowledge, and demonstrated a significant reduction, up to an order or two of magnitude, in the number of TIRPs discovered when using it, as well as in the runtime needed to extract these TIRPs. Nevertheless, this reduced set of TIRPs, when serving as features for classification and prediction, using any of four different families of classifier-induction algorithms, in three different clinical domains, proved to be as good as the whole set with respect to classification and prediction performance. Overall, the CSAC version, the most restrictive of the SAC versions, seemed to be the most promising for inducing the smallest set of TIRPs, while maintaining the same classification and prediction performance.

**Acknowledgments**: The authors wish to thank all their clinical collaborators for assisting in developing the clinical knowledge bases. A. Shknevsky and Y. Shahar were partially supported by the MobiGuide project, partially funded by the European Commission 7th Framework Programme grant #287811.

# Appendix A

**The SAC pseudo-Code within the KarmaLego Algorithm**

The KarmaLego algorithm (Algorithm 1) consists of two main phases [Moskovitch and Shahar, 2015a]. The first phase is called Karma and it enumerates all 2-sized TIRPs (Algorithm 2). The second phase is called Lego and it is a recursive method that extends each k-sized TIRP into the possible (k+1)-sized TIRPs (Algorithm 3).

Note that the supplied algorithms are coherent with Moskovitch and Shahar's original paper, except the underlined modifications to support the SAC principle. Additional implementation details can be found in Shknevsky's M.Sc. thesis [Shknevsky, 2014].

Algorithm 1 – The KarmaLego main loop

**Input:**

db - A database of |E| entities.

min_ver_sup - The minimal vertical support threshold.

**Output:** T - An enumerated tree of all frequent TIRPs.

**Algorithm:**

1. T = Karma(db, min_ver_support)
2. Foreach $t \in T^2$ // $T^2$ is T at the $2^{nd}$ level
2.1. Lego(T, t, min_ver_support)
3. Return T

Algorithm 2 – The Karma algorithm with the SAC modifications

**Input:**

db - A database of |E| entities, representing for each entity $e \in E$ the lexicographically sorted vector of its symbolic time intervals, $e.I$

min_ver_sup - The minimal vertical support threshold.

**Output:** T - An enumerated tree of up to 2-sized frequent TIRPs.

**Algorithm:**

1. T ← ∅
2. Foreach $e \in E$ // $T^2$ is T at the $2^{nd}$ level
2.1. Foreach $I^i, I^j \in e.I \wedge i < j$
2.1.1. r ← the temporal relation among $I^i, I^j$
2.1.2. <u>if SSAC ∨ (LSAC ∧ $e.I^i_{sym} \neq e.I^j_{sym}$) ∨ CSAC</u>
2.1.2.1. <u>Foreach $I^k \in e.I \wedge k \neq i \wedge k < j$</u>
2.1.2.1.1. <u>If $I^k.e > I^i.e \wedge$
   $(sem\_type(e.I^i) = sem\_type(e.I^k) \vee sem\_type(e.I^k) = sem\_type(e.I^j))$</u>
2.1.2.1.1.1. <u>break</u>
2.1.3. Index($T^2, < e.I^i_{sym}, r, e.I^j_{sym} >$
3. Foreach $t \in T^2$
3.1. If ver_sup(t) < min_ver_sup
3.1.1. Prune(t)
4. Return T

Note that regarding the use of the sem_type method, we saved for each symbolic time interval (STI), its symbol (sym) as a pair of concept and value. In this case, checking the semantic equivalence of symbolic intervals is interpreted as a comparison of the concepts (and not the values), corresponding to Definition 1.

Algorithm 3 – The Lego algorithm

**Input:**

T – The enumeration tree created by Karma.

t – A TIRP that has to be extended.

min_ver_sup - The minimal vertical support threshold.

**Output:** void.

**Algorithm:**

1. Foreach $sym \in T^1$

1.1. Foreach $r \in R$

1.1.1. Create new $t^c$ of size (t.size +1)

1.1.2. $t^c.s[t^c.size - 1] \leftarrow sym$

1.1.3. $t^c.tr[t^c.tr\_size - 1] \leftarrow r$

1.1.4. $C \leftarrow 0$

1.1.5. $C \leftarrow generate\ cantidate\ TIRPs\ from\ t^c$

1.1.6. Foreach $c \in C$ // candidates

1.1.6.1. Search_Supporting_Instances(c, $T^2$)

1.1.6.1.1. If ver_sup(c) > min_ver_sup

1.1.6.1.1.1. $T \leftarrow T \cup c$ // c is frequest

1.1.6.1.1.2. Lego(T, c, min_ver_sup)

2. Return T

The Search_Supporting_Instances method (Algorithm 4) receives as input the candidate TIRP c and the set of the two sized TIRPs in T2. In line 1, the next (new) symbol that was added (in Algorithm 3) is set to next_sym; then, for each instance in the extended TIRP's supporting instances, the search is made. First the temporal relation rel between the next time interval and the latest (in the extended TIRP) is set (line 2.1), then the latest symbol of the extended TIRP sym is set (line 2.2).

GetNextSTIs (Line 2.3) searches the symbolic time interval (sti) two-dimensional square array $T^2$, using indices defined by the symbols sym and last_sym and the temporal relation rel, for the instances starting with the latest time interval in the instance ins.sti. GetNextSTIs might return several new symbolic time intervals, next_stis.

The method $NoInst?$ searches for an instance of a pair of symbolic time intervals having the given temporal relation; thus, it gets sym, rel, and next_sym as indices to its fourth argument, the appropriate $T^2$ array entry, in which it queries the HashMap for the pair based on the entity id of the new instance ($inst^{new}.e\_id$) and the first symbolic time interval instance ($inst^{new}.sti[c.size - 2 - i]$). It returns True if no instance of the relation was found; else, it returns False.

The method $SAC?$ (e_id, sti_1, sti_2, $T^2[sti\_1.sym, rel, sti\_2.sym]$) checks the relation between a given pair of symbolic time intervals, sti_1 and sti_2, for the entity e_id in the $T^2$ the SAC compatibility. If we are checking without SAC, then the result is True. If we are checking using LSAC, but sem_type(sti_1) = sem_type(sti_2) then the result is True. Otherwise (i.e., when using SSAC, CSAC, or LSAC but sem_type(sti_1) ≠ sem_type(sti_2)), then, if there is a gap between sti_1 and sti_2, then $SAC?$ will return True if $T^2[sti\_1.sym, rel, sti\_2.sym]$ contains the pair (sti_1, sti_2), which means that it was previously discovered as obeying the SAC criterion.

Algorithm 4 – Search_Supporting_Instances

**Input:**

c – TIRP to extend by searching supporting symbolic intervals instances.

$T^2$ – A the 2-dimential array of 2-sized TIRPs instances.

min_ver_sup - The minimal vertical support threshold.

**Output:** c – extended by the supporting instances.

**Algorithm:**

1. $next\_sym \leftarrow c.sym[c.size - 1]$

2. Foreach inst $\in c.insts$

2.1. rel ← c.tr[c.tr_size-1]

2.2. sym ← c.sym[c.size-2]

2.3. next_stis ← GetNextSTIs(inst.sti[c.size-2], $T^2$[sym, rel, next_sym])

2.4. Foreach next_sti ∈ next_stis

2.4.1. $inst^{new} \leftarrow inst$

2.4.2. <u>if ¬NoSAC ∧ ¬SAC? $(inst^{new}.e\_id, inst^{new}.sti[c.size - 2],$
              $next\_sti, T^2[sym, rel, next\_sym])$</u>

2.4.2.1. <u>break</u>

2.4.3. For (i=1; i<c.size-1; i++)

2.4.3.1. rel ← c.tr[c.tr_size-1-i]

2.4.3.2. sym ← c.sym[c.size-2-i]

2.4.3.3. <u>if NoInst? $(inst^{new}.e\_id, inst^{new}.sti[c.size - 2 - i],$
              $next\_sti, T^2[sym, rel, next\_sym])$</u>

2.4.3.3.1. <u>break</u>

2.4.3.4. <u>if ¬NoSAC ∧ ¬SSAC ∧
              ¬SAC? $(inst^{new}.e\_id, inst^{new}.sti[c.size - 2 - i],$
              $next\_sti, T^2[sym, rel, next\_sym])$</u>

2.4.3.4.1. <u>break</u>

2.4.4. c.insts ← c.insts ∪ $insts^{new}$

2.5. remove inst from c.insts

3. Return c

# Appendix B

**Data sets and Knowledge Base definitions**

We describe here the data sets used in our experiments, and the definitions we used in the case of the knowledge-based temporal abstraction method.

**The Oncology dataset**

A medical oncology domain knowledge source that was used for the evaluation (of the KB temporal abstraction method) was an oncology knowledge data base specific to the bone-marrow transplantation domain. It includes in total more than 350 concepts from 1991 to 1997: more than 200 raw concepts (e.g., laboratory tests – White Blood Cell count, Hemoglobin), internal events (e.g., bone marrow transplantations – BMT), and more. The data source used for the evaluation was of bone-marrow transplantation patients who were followed for 2-4 years at the Rush Medical Center, Chicago, USA. The knowledge and data bases were previously elicited from our clinicians' colleagues. Table 2 presents the knowledge that was used for the purposes of this study. Note that in case of an overlap the maximum value will be taken.

We used 207 patients who had a bone marrow transplantation and data for the following 12 laboratory tests: Platelet count, Hemoglobin, White Blood Cell count (WBC), Glucose levels, Total Bilirubin, Alkaline Phosphatase, Hematocrit, Monocytes, Lymphocytes, Eosinophil granulocyte count (EOS), Neutrophilic band forms (Bands), Basophil granulocyte count (Basos).

Table 2 - The knowledge base for the oncology dataset.

| Platelet count | | Hemoglobin | | WBC | |
|---|---|---|---|---|---|
| High | ≥400 | High | ≥16 | Very_High | ≥20 |
| Normal | 100-400 | Normal | 11-16 | High | 12-20 |
| Moderately_Lo | 50-100 | Moderately_Lo | 9-11 | Normal | 2.5-12 |
| Low | 20-50 | Low | 7-9 | Moderately_Low | 0.5-2.5 |
| Very_Low | <20 | Very_Low | <7 | Low | 0.1-0.5 |
| | | | | Very_Low | <0.1 |
| **Glucose level** | | **Total Bilirubin** | | **Alkaline Phosphatase** | |
| Very_High | ≥250 | Very_High | ≥10 | Very_High | ≥225 |
| High | 151-250 | High | 3-10 | High | 110-225 |
| Normal | 75-151 | Normal | 1.5-3 | Normal | 35-110 |
| Low | <75 | Low | <1.5 | Low | <35 |
| **Hematocrit** | | **Monocytes** | | **Lymphocytes** | |
| High | ≥46.9 | High | ≥10 | High | ≥52 |
| Normal | 34.9-46.9 | Normal | 3-10 | Normal | 18-52 |
| Low | <34.9 | Low | <3 | Low | <18 |
| **EOS** | | **Bands** | | **Basos** | |
| Very_High | ≥12 | High | >=6 | High | >=3 |
| High | 6-12 | Normal | <6 | Normal | <3 |
| Normal | <6 | | | | |

For the interpolation, for producing intervals out of the time-stamped raw data, we treated each time-stamped point as good for one day after and one day before. The task was to classify patients who went through autologous bone-marrow transplantation (137 patients) versus allogeneic bone-marrow transplantation (70 patients) based on TIRPs discovered from the mentioned laboratory tests.

**The Hepatitis dataset**

The hepatitis dataset contains the results of laboratory examinations on hepatitis B and C patients who were admitted to Chiba University Hospital in Japan. Hepatitis A, B, and C are viral infections that affect the liver of the patient. Hepatitis B and C chronically inflame the hepatocytes, whereas hepatitis A acutely inflames them. Hepatitis B and C are especially important because they have a potential risk for developing liver cirrhosis or hepatocarcinoma. The dataset contains long time-series data of laboratory examinations. The subjects are 771 patients with hepatitis B and C who were examined between 1982 and 2001. Table 3 presents the relevant knowledge that was extracted from a public KDD conference challenge [Ho and Nguyen, 2003] and was used for our evaluation; In case of an overlap the maximum value was taken. The data set is **publicly** available [Berka et al., 2002].

We used 499 patients who had a biopsy result of hepatitis B (204 patients) or C (295 patients) and the ten most frequent tests (occurring in most of the patients), including: Glutamic-Oxaloacetic Transaminase (GOT), Glutamic-Pyruvic Transaminase (GPT), Lactate DeHydrogenase (LDH), Total Protein (TP), ALkaline Phosphatase (ALP), Albumin (ALB), Uric Acid (UA), Total BILirubin (T-BIL), Indirect BILirubin (I-BIL), and Direct BILirubin (D-BIL). For the interpolation, we treated each time-stamped point as good for 15 days before and after each point. The task was to classify the patients as Hepatitis B versus Hepatitis C, based on the TIRPs discovered from the mentioned 10 most frequent tests.

Table 3 - The knowledge base for the hepatitis dataset.

| GOT    |        | GPT    |        | LDH    |         |
|--------|--------|--------|--------|--------|---------|
| High   | ≥40    | High   | ≥40    | High   | ≥450    |
| Normal | 7-40   | Normal | 7-40   | Normal | 216-450 |
| Low    | <7     | Low    | <7     | Low    | <216    |
| **TP**     |        | **ALP**    |        | **ALB**    |         |
| High   | ≥8.2   | High   | ≥206   | High   | ≥5.1    |
| Normal | 6.5-8.2| Normal | 72-206 | Normal | 3.9-5.1 |
| Low    | <6.5   | Low    | <72    | Low    | <3.9    |
| **UA**     |        | **T-BIL**  |        | **I-BIL**  |         |
| High   | ≥8     | High   | ≥1.2   | High   | ≥0.9    |
| Normal | 2.5-8  | Normal | 0.2-1.2| Normal | 0.2-0.9 |
| Low    | <2.5   | Low    | <0.2   | Low    | <0.2    |
| **D-BIL**  |        |        |        |        |         |
| High   | ≥3     |        |        |        |         |
| Normal | <3     |        |        |        |         |

**The Diabetes dataset**

The diabetes data set was provided by the National Institute for Biotechnology in the Negev (NIBN) in a joint study with Soroka University Medical Center [Gordon, 2012]. The subjects are 26k anonymous patients (and about 12 million raw data records) who had diabetes and various laboratory tests between 2004 and 2008. The data include static information (e.g., gender) and temporal records (e.g., High-density lipoprotein, Low-density lipoprotein, Triglycerides, Glucose, Hemoglobin A1c, Creatinine, Total cholesterol, and Albuminuria). The main interest in this data was on the investigation of factors associated with changes in renal function (mostly focusing on the level of albuminuria, or secretion of protein in the urine), exploring its predictive risk factors.

We used 5178 patients who had Albumin-to-creatinine ratio or Albumin-24 hours from urine in the last, fifth, year of the data set, and who also had these tests and also Glycosylated hemoglobin (HbA1c) and Creatinine (CREATININE) in the first four years of the data set. For the interpolation, we treated each time-stamped point of Albuminuria ACR and Albuminuria U24h as good for 3 months before and after each point. For Creatinine as good for 2 months before and after each point. And for HBA1C as good for 4 months before and after each point. The task was to predict Albuminuria-normo (3231 patients) versus micro- or macro-albuminuria (1947 patients) in the fifth year based on TIRPs discovered in the first four years. Table 4 presents the relevant knowledge that was supplied by Gordon [Gordon, 2012] and other clinicians who worked on other projects as well.

Table 4 - The knowledge base for the diabetes dataset.

| **Albuminuria ACR/Albuminuria U24h** | | | | | | |
|---|---|---|---|---|---|---|
| Female | Macro | >300 | Male | Macro | >300 |
| | Micro | 30-300 | | Micro | 30-300 |
| | Normo-High | 15-30 | | Normo-High | 13-30 |
| | Normo-Low | 0-15 | | Normo-Low | 0-13 |
| **CREATININE** | | | | | | |
| Female | Very_High | >4 | Male | Very_High | >4 |
| | High | 2-4 | | High | 2-4 |
| | Moderately_High | 1-2 | | Moderately_High | 1.2-2 |
| | Normal | <1 | | Normal | <1.2 |
| **HbA1c** | | | | | | |
| Very_High | >10.5 | | | | |
| High | 9-10.5 | | | | |
| Moderately_High | 7-9 | | | | |
| Normal | <7 | | | | |